%% file: main.tex
\documentclass[sigconf, nonacm]{acmart}

\usepackage{booktabs}
\usepackage{enumitem}
\usepackage{tabularx}
\usepackage{tikz}
\usetikzlibrary{positioning, arrows.meta, calc, shapes.geometric, shapes.misc, fit}

\AtBeginDocument{%
  \emergencystretch=1.5em\relax
  \setlength{\textfloatsep}{8pt plus 2pt minus 2pt}%
  \setlength{\floatsep}{6pt plus 2pt minus 2pt}%
  \setlength{\intextsep}{6pt plus 2pt minus 2pt}%
  \setlength{\dbltextfloatsep}{10pt plus 2pt minus 2pt}%
  \setlength{\dblfloatsep}{6pt plus 2pt minus 2pt}%
  \captionsetup{skip=6pt}%
  \AtEndEnvironment{table}{\par\vskip-4pt}%
  \AtEndEnvironment{table*}{\par\vskip-4pt}%
  \AtEndEnvironment{figure}{\par\vskip-4pt}%
  \AtEndEnvironment{figure*}{\par\vskip-4pt}%
  \widowpenalty=0 \clubpenalty=0 \brokenpenalty=0 }

\newcommand{\rev}[1]{#1}
\definecolor{revgreen}{RGB}{0,110,40}
\newcommand{\revg}[1]{#1}

\input{results/numbers-part1}

\input{results/numbers-part2}

\setcopyright{none}
\title[What Does an Ontology Catch?]{What Does an Ontology Catch? Curated, LLM-Generated and Data-Derived Ontologies as Checkers of LLM Agents over Knowledge Graphs}

\author{Michal Orzechowski}
\orcid{0000-0002-8558-1283}
\affiliation{%
  \institution{ACK Cyfronet AGH}
  \city{Krakow}
  \country{Poland}}
\affiliation{%
  \institution{AGH University of Krakow}
  \city{Krakow}
  \country{Poland}}
\affiliation{%
  \institution{Sano Centre for Computational Medicine}
  \city{Krakow}
  \country{Poland}}

\author{Rund Tawfiq}
\orcid{0000-0002-7052-573X}
\affiliation{%
  \institution{Sano Centre for Computational Medicine}
  \city{Krakow}
  \country{Poland}}

\author{Tomasz Kosciolek}
\orcid{0000-0002-9915-7387}
\affiliation{%
  \institution{Sano Centre for Computational Medicine}
  \city{Krakow}
  \country{Poland}}
\affiliation{%
  \institution{AGH University of Krakow}
  \city{Krakow}
  \country{Poland}}

\begin{document}

\begin{abstract}
LLM agents that answer questions over Web knowledge graphs are increasingly
wrapped in checks derived from the graph's ontology, on the assumption that a
formal schema makes a reliable judge. Yet a checker can be wrong as well: it can
miss errors, and it can flag correct answers and push the agent into ``fixing''
them. Whether a curated ontology is worth its cost as such a judge, and how it
compares with an ontology an LLM writes or a schema mined from the data, is
unknown. We make the checker itself the object of study. Across curated,
LLM-generated and data-derived ontologies, used as prompt context, as a
checker, or both, and across open-weight and proprietary agents on public
graphs and a private graph of nearly $10^{9}$ triples, we account for what
ontology-derived checks catch and what they wrongly flag. We show how each
source trades caught errors against false alarms, how a curated schema used as
published can contradict its own data, which errors structural checks
miss, and whether combining sources or giving the agent the checker's specific
feedback pays off. \revg{On UniProt, the curated ontology flags more correct answers than either an ontology derived from the data or one an LLM derives from a summary of the data with masked identifiers. On real agent answers, its flags are only marginally more precise than random flags.}
\end{abstract}


\maketitle

\input{sections/introduction}
\input{sections/related-work}
\input{sections/checks}
\input{sections/design}
\input{sections/results}
\input{sections/discussion}
\input{sections/conclusion}

\begin{acks}
This work is co-financed by the project of the Minister of Science and Higher Education (MEiN/2023/DIR/3796) and by the subvention assigned to AGH University of Krakow. We acknowledge ACK Cyfronet AGH (PLGINT/2026/020157, Gaia/AIF/101314359, LAIF/AIF/101234208) for providing computational resources. T.K. and R.T. are supported by the National Science Centre, Poland grant 2023/05/Y/NZ2/00080.

\vspace{\baselineskip}\noindent
Large language models are the subject of this study: they serve as the agents and the LLM judge, and they write the LLM-generated ontologies (Sections 3 and 4). They were also used as tools for experiment orchestration and design, data analysis, and copy-editing of the manuscript. All content was reviewed and validated by the authors.
\end{acks}

\bibliographystyle{ACM-Reference-Format}
\bibliography{references}

\appendix
\input{sections/appendix}

\end{document}

%% file: results/numbers-part1.tex
\newcommand{\resConfCFiveClaudeHaikuFiveFiveEst}{1.0}

\newcommand{\resConfCFiveDeepseekAiDeepSeekVFourFlashZeroSevenThreeOneEst}{2.0}

\newcommand{\resConfCFiveGoogleGemmaFourThreeOneBEst}{6.2}

\newcommand{\resConfCFiveP}{1.000}
\newcommand{\resConfCFivePBootstrap}{0.457}

\newcommand{\resConfCFivePooledOverPEst}{0.2}
\newcommand{\resConfCFivePooledOverPHi}{3.1}
\newcommand{\resConfCFivePooledOverPLo}{\ensuremath{-}3.2}

\newcommand{\resConfCFiveQwenEst}{0.0}
\newcommand{\resConfCFiveQwenQwenThreeCoderThreeZeroBAThreeBInstructEst}{\ensuremath{-}7.1}

\newcommand{\resConfCFiveSonnetEst}{0.9}

\newcommand{\resConfCFiveZaiOrgGLMFiveThreeFlashEst}{\ensuremath{-}1.8}

\newcommand{\resConfCFourP}{1.000}
\newcommand{\resConfCFourPBootstrap}{1.000}

\newcommand{\resConfCFourUnionCensusEst}{\ensuremath{-}10.6}
\newcommand{\resConfCFourUnionCensusHi}{\ensuremath{-}9.2}
\newcommand{\resConfCFourUnionCensusLo}{\ensuremath{-}12.2}

\newcommand{\resConfCFourUnionCuratedEst}{\ensuremath{-}0.5}
\newcommand{\resConfCFourUnionCuratedHi}{0.4}
\newcommand{\resConfCFourUnionCuratedLo}{\ensuremath{-}1.5}

\newcommand{\resConfCOneAP}{0.003}
\newcommand{\resConfCOneAPBootstrap}{0.002}

\newcommand{\resConfCOneAPooledOverPEst}{\ensuremath{-}2.1}
\newcommand{\resConfCOneAPooledOverPHi}{\ensuremath{-}0.9}
\newcommand{\resConfCOneAPooledOverPLo}{\ensuremath{-}3.4}

\newcommand{\resConfCOneBP}{1.000}
\newcommand{\resConfCOneBPBootstrap}{0.113}

\newcommand{\resConfCOneBQwenEst}{\ensuremath{-}1.1}
\newcommand{\resConfCOneBQwenHi}{1.9}
\newcommand{\resConfCOneBQwenLo}{\ensuremath{-}4.2}

\newcommand{\resConfCOneBSonnetEst}{\ensuremath{-}0.8}
\newcommand{\resConfCOneBSonnetHi}{0.9}
\newcommand{\resConfCOneBSonnetLo}{\ensuremath{-}2.7}

\newcommand{\resConfCThreeCuratedMeanMaskedEst}{15.7}
\newcommand{\resConfCThreeCuratedMeanMaskedHi}{17.9}
\newcommand{\resConfCThreeCuratedMeanMaskedLo}{13.7}

\newcommand{\resConfCThreeP}{0.025}
\newcommand{\resConfCThreePBootstrap}{\ensuremath{<}0.001}

\newcommand{\resConfCTwoP}{0.013}
\newcommand{\resConfCTwoPBootstrap}{\ensuremath{<}0.001}

\newcommand{\resConfCTwoTypeWrongRecallCuratedCensusEst}{\ensuremath{-}22.3}
\newcommand{\resConfCTwoTypeWrongRecallCuratedCensusHi}{\ensuremath{-}21.0}
\newcommand{\resConfCTwoTypeWrongRecallCuratedCensusLo}{\ensuremath{-}23.5}

\newcommand{\resLedgerUniprotNoneCheckerCuratedBankFalseFlagRate}{15.0}

\newcommand{\resLedgerUniprotNoneCheckerCuratedBankNonEmptyRecall}{54.3}

\newcommand{\resLedgerUniprotNoneCheckerCuratedBankRecall}{60.1}

\newcommand{\resLedgerUniprotNoneCheckerCuratedUnknownRate}{1.8}

\newcommand{\resLedgerUniprotNoneCheckerGenericUnknownRate}{1.8}

\newcommand{\resLedgerUniprotNoneReplayCensusBankFalseFlagRate}{9.7}

\newcommand{\resLedgerUniprotNoneReplayCensusBankNonEmptyRecall}{62.1}

\newcommand{\resLedgerUniprotNoneReplayCensusBankRecall}{75.6}

\newcommand{\resLedgerUniprotNoneReplayCensusUnknownRate}{1.6}

\newcommand{\resLedgerUniprotNoneReplayGeneratedDcOneBankFalseFlagRate}{1.2}

\newcommand{\resLedgerUniprotNoneReplayGeneratedDcOneBankNonEmptyRecall}{47.2}

\newcommand{\resLedgerUniprotNoneReplayGeneratedDcOneBankRecall}{55.2}

\newcommand{\resLedgerUniprotNoneReplayGeneratedDcOneUnknownRate}{2.0}

\newcommand{\resLedgerUniprotNoneReplayGeneratedDcTwoBankFalseFlagRate}{1.3}

\newcommand{\resLedgerUniprotNoneReplayGeneratedDcTwoBankNonEmptyRecall}{48.5}

\newcommand{\resLedgerUniprotNoneReplayGeneratedDcTwoBankRecall}{56.4}

\newcommand{\resLedgerUniprotNoneReplayGeneratedOneUnknownRate}{2.0}

\newcommand{\resLedgerUniprotNoneReplayGeneratedThreeUnknownRate}{1.5}

\newcommand{\resLedgerUniprotNoneReplayGeneratedTwoUnknownRate}{1.9}

\newcommand{\resLedgerUniprotNoneReplayUnionUnknownRate}{1.4}

\newcommand{\resUniprotClaudeHaikuFiveFiveCensusPromptAccuracy}{93.0}

\newcommand{\resUniprotClaudeHaikuFiveFiveCuratedPromptAccuracy}{93.8}

\newcommand{\resUniprotClaudeHaikuFiveFiveNoneCheckerCuratedNetChange}{0.2}

\newcommand{\resUniprotClaudeHaikuFiveFiveNonePromptAccuracy}{95.7}

\newcommand{\resUniprotDeepseekAiDeepSeekVFourFlashZeroSevenThreeOneCensusPromptAccuracy}{86.2}

\newcommand{\resUniprotDeepseekAiDeepSeekVFourFlashZeroSevenThreeOneCuratedPromptAccuracy}{84.8}

\newcommand{\resUniprotDeepseekAiDeepSeekVFourFlashZeroSevenThreeOneNoneCheckerCuratedNetChange}{1.0}

\newcommand{\resUniprotDeepseekAiDeepSeekVFourFlashZeroSevenThreeOneNonePromptAccuracy}{82.2}

\newcommand{\resUniprotGoogleGemmaFourThreeOneBCensusPromptAccuracy}{52.7}

\newcommand{\resUniprotGoogleGemmaFourThreeOneBCuratedPromptAccuracy}{44.4}

\newcommand{\resUniprotGoogleGemmaFourThreeOneBNoneCheckerCuratedNetChange}{1.0}

\newcommand{\resUniprotGoogleGemmaFourThreeOneBNonePromptAccuracy}{15.7}

\newcommand{\resUniprotQwenCensusPromptAccuracy}{80.8}

\newcommand{\resUniprotQwenCuratedPromptAccuracy}{79.7}

\newcommand{\resUniprotQwenGeneratedOnePromptAccuracy}{81.9}

\newcommand{\resUniprotQwenNoneCheckerCuratedNetChange}{0.8}

\newcommand{\resUniprotQwenNonePromptAccuracy}{80.8}

\newcommand{\resUniprotQwenQwenThreeCoderThreeZeroBAThreeBInstructCensusPromptAccuracy}{9.0}

\newcommand{\resUniprotQwenQwenThreeCoderThreeZeroBAThreeBInstructCuratedPromptAccuracy}{5.9}

\newcommand{\resUniprotQwenQwenThreeCoderThreeZeroBAThreeBInstructNoneCheckerCuratedNetChange}{\ensuremath{-}0.2}

\newcommand{\resUniprotQwenQwenThreeCoderThreeZeroBAThreeBInstructNonePromptAccuracy}{1.6}

\newcommand{\resUniprotSonnetCensusPromptAccuracy}{96.0}

\newcommand{\resUniprotSonnetCuratedPromptAccuracy}{95.2}

\newcommand{\resUniprotSonnetGeneratedOnePromptAccuracy}{97.5}

\newcommand{\resUniprotSonnetNoneCheckerCuratedNetChange}{0.3}

\newcommand{\resUniprotSonnetNonePromptAccuracy}{95.6}

\newcommand{\resUniprotZaiOrgGLMFiveThreeFlashCensusPromptAccuracy}{87.1}

\newcommand{\resUniprotZaiOrgGLMFiveThreeFlashCuratedPromptAccuracy}{86.5}

\newcommand{\resUniprotZaiOrgGLMFiveThreeFlashNoneCheckerCuratedNetChange}{0.6}

\newcommand{\resUniprotZaiOrgGLMFiveThreeFlashNonePromptAccuracy}{82.4}

%% file: results/numbers-part2.tex
\newcommand{\resUniprotBankBankReplayCensusPrecisionAtPrevalenceTwoFive}{72.3}

\newcommand{\resUniprotBankBankReplayCuratedPrecisionAtPrevalenceTwoFive}{57.1}

\newcommand{\resUniprotBankBankReplayGeneratedDcOnePrecisionAtPrevalenceTwoFive}{93.7}

\newcommand{\resUniprotBankBankReplayGeneratedDcTwoPrecisionAtPrevalenceTwoFive}{93.4}

%% file: sections/introduction.tex
\section{Introduction}
\label{sec:introduction}
Public data across domains is increasingly published as knowledge graphs, which describe entities and the relationships between them so that a single query can traverse many entities; Wikidata \cite{vrandecic2014}, DBpedia \cite{lehmann2015} and UniProt \cite{uniprot2025}, for example, can be queried through SPARQL endpoints \cite{arenas2011}. Most graphs are described by an ontology, a machine-readable account of what kinds of entities a graph contains and how they are related.

\begin{figure*}[t]
  \centering
  \resizebox{0.85\textwidth}{!}{\input{figures/overview.tex}}
  \Description{A flow diagram: a question goes to an LLM agent, which sends a SPARQL query to a knowledge graph; a check derived from an ontology reads the query and its results and either passes the answer or flags it, which sends the agent back for one retry. Below, an ontology from one of three sources (curated, written by an LLM, derived from the data) feeds the agent as context and the check.}
  \caption{The setting. An LLM agent answers a question with a SPARQL query; a check derived from an ontology passes the answer or flags it, and a flag triggers one retry. The same ontology can also be the agent's context.}
  \label{fig:overview}
\end{figure*}
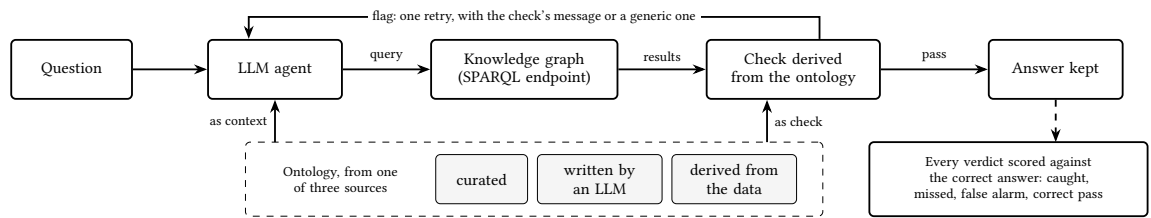

Large language models (LLMs) have made knowledge graphs more accessible, as they translate natural language questions into SPARQL queries, run them against the endpoints, and report the retrieved results \cite{liu2024,smeros2026,kinjo2026}. However, it is difficult to validate their results. A query can be syntactically valid, execute without error, yet answer a different question from the one asked \cite{brass2006,liu2025}. One way to assess an answer is to test whether its query fits what the graph's ontology states about its entities and relationships. If the ontology contains, in the Web Ontology Language (OWL) \cite{owl2012}, a domain axiom stating that whatever has a birth date is a person and a disjointness axiom stating that no city is a person, a query for the birth date of a city contradicts it. These kinds of checks were built for SPARQL queries written by LLMs, where they are reported to increase accuracy \cite{allemang2024}, and testing a graph's data against its ontology with SPARQL is an established method to assess the quality of Linked Data \cite{kontokostas2014}.

Checks may still permit errors. They may accept incorrect answers or flag correct ones, prompting an LLM agent to revise a correct answer into an incorrect one, as has been observed in text-to-SQL refinement \cite{hong2026}. Accuracy alone reveals neither error. Checks based on ontologies are particularly prone to flagging correct answers, because axioms are read as constraints on the data \cite{motik2009} while published ontologies often do not describe all of it: in a crawl of Linked Data, 14.3\% of triples used properties that were not defined in any of the crawled vocabularies \cite{hogan2010}, and a check that requires every property to be declared would reject correct answers that rely on them. Before an ontology can be used to judge an agent's answer, its suitability for that role must itself be assessed.

Existing approaches to ontology evaluation do not address this use. Ontologies
are typically evaluated by comparison with a gold standard, by expert
assessment, or by their coverage of a corpus or dataset \cite{brank2005}.
Constraints derived from ontologies were also compared with constraints mined
from the data to validate the data \cite{duan2024}, where a constraint
violation is interpreted as an error in the data. When an ontology is used to
check an agent's answers, however, a violation may instead reflect an error in
the ontology, resulting in a correct answer being rejected. We therefore
evaluate ontologies in this application directly, measuring both the errors
they detect and the correct answers they incorrectly flag.

This gap has implications for both knowledge graph publishers and agent developers. Publishers would benefit from knowing how expert curated ontologies compare with ontologies generated by LLMs \cite{zhang2025}, or ontologies derived from data \cite{rabbani2023} when used to assess agent answers. Developers may benefit from understanding whether an ontology is most useful as context for generating an answer, as a check on the answer, or both.

In this paper, we evaluate curated, LLM-generated, and data-derived ontologies as checks on agents' answers. We address three research questions:
\begin{description}[leftmargin=2.6em,labelwidth=2.2em,labelsep=0.4em,topsep=1pt,partopsep=0pt]
  \item[RQ1] Which errors do checks from curated, LLM-generated and
    data-derived ontologies catch, how often do they flag correct answers, and
    what do they cost to set up and to run?
  \item[RQ2] What causes the false alarms, and which errors do the checks miss?
  \item[RQ3] How does the ontology contribute at other stages of the process: as context for answer generation, in combination with another check, or as feedback to the agent after a check is triggered?
\end{description}

A fair comparison must address three concerns, and each shapes our design. First, executing a query can itself reveal errors, for example when it returns no results, so we distinguish errors caught by a check from those already exposed by query execution. Second, results on artificially introduced faults may not generalize to real errors \cite{just2014,pearson2017}, so we evaluate each check on both real agent answers and faults introduced into correct queries. Third, LLMs may recall identifiers from public knowledge graphs \cite{gashkov2025}, potentially biasing the ontologies they generate. We therefore also have LLMs build ontologies from a masked summary of the graph, in which identifiers are replaced with opaque names. Every check judges the same answers and faults, and each decision is compared with the pre-defined answer (Figure~\ref{fig:overview}).

This paper makes three contributions:
\begin{itemize}
  \item A comparison of curated, LLM-generated and data-derived ontologies as
    checks on LLM agents built on five open-weight and two proprietary models,
    confirmatory on UniProt and exploratory on other public graphs and a
    private graph of nearly $10^9$ triples.
  \item An analysis of why such checks go wrong: where a curated ontology as
    published disagrees with its own data, and which errors the checks
    miss.
  \item A released set of artificial faults introduced into correct queries,
    with their pre-defined correct answers, and a protocol that measures any check's catches and false alarms
    against a retry on empty results, without running an agent.\footnote{The
    artificial faults, the code and the pre-registration are available at
    \url{https://anonymous.4open.science/r/www-2027-replication-package-36D6}; Appendix~\ref{sec:app-package} lists the contents.}
\end{itemize}

%% file: figures/overview.tex
\begin{tikzpicture}[
    every node/.append style={font=\footnotesize},
    box/.style={draw, rounded corners=2pt, thick, minimum height=0.9cm,
                align=center, inner sep=2pt},
    src/.style={draw, rounded corners=2pt, minimum height=0.75cm,
                align=center, inner sep=1pt, fill=black!4},
    arrow/.style={-{Stealth[length=2mm]}, thick},
    lab/.style={font=\scriptsize}
  ]
  \node[box, minimum width=1.8cm, text width=1.7cm] (Q)   at ( 0.0, 0) {Question};
  \node[box, minimum width=2.0cm, text width=1.9cm] (A)   at ( 3.1, 0) {LLM agent};
  \node[box, minimum width=2.8cm, text width=2.7cm] (KG)  at ( 6.9, 0) {Knowledge graph\\(SPARQL endpoint)};
  \node[box, minimum width=2.6cm, text width=2.5cm] (C)   at (11.0, 0) {Check derived\\from the ontology};
  \node[box, minimum width=2.0cm, text width=1.9cm] (OUT) at (15.0, 0) {Answer kept};

  \draw[arrow] (Q)  -- (A);
  \draw[arrow] (A)  -- (KG)  node[midway, above, lab] (Lq) {query};
  \draw[arrow] (KG) -- (C)   node[midway, above, lab] (Lr) {results};
  \draw[arrow] (C)  -- (OUT) node[midway, above, lab] (Lp) {pass};

  \draw[arrow] (11.4, 0.465) -- (11.4, 0.8) -- (2.7, 0.8) -- (2.7, 0.465);
  \node[lab, fill=white, inner sep=2pt, anchor=center] (Lf) at (7.05, 0.8)
        {flag: one retry, with the check's message or a generic one};

  \node[draw, dashed, rounded corners=2pt, minimum width=8.8cm, minimum height=1.13cm]
        (BAND) at (7.05, -1.675) {};
  \node[lab, text width=2.55cm, align=center, inner sep=1pt] (SRCT) at (4.1, -1.675)
        {Ontology, from one\\of three sources};
  \node[src, minimum width=1.4cm, text width=1.3cm] (S1) at (6.25, -1.675) {curated};
  \node[src, minimum width=1.9cm, text width=1.8cm] (S2) at (8.05, -1.675) {written by an LLM};
  \node[src, minimum width=1.9cm, text width=1.8cm] (S3) at (10.1, -1.675) {derived from the data};

  \draw[arrow] (3.1, -1.11)  -- (A.south) node[midway, left,  lab] (Lc) {as context};
  \draw[arrow] (10.6, -1.11) -- (10.6, -0.465) node[midway, right, lab] (Lk) {as check};

  \node[box, minimum width=4.24cm, text width=4.1cm, minimum height=1.13cm, font=\scriptsize] (SCORE)
        at (14.28, -1.675)
        {Every verdict scored against\\the correct answer: caught,\\missed, false alarm, correct pass};
  \draw[arrow, dashed] (OUT.south) -- (15.0, -1.11);
\end{tikzpicture}

%% file: sections/related-work.tex
\section{Related Work}
\label{sec:related-work}

LLM agents are increasingly used to interact with knowledge graphs. TogoMCP makes approximately sixty RDF databases accessible to LLMs through the Model Context Protocol (MCP) \cite{kinjo2026}. Other systems focus on generating and executing queries. SPINACH generates SPARQL queries and revises them based on returned results, empty results, and execution errors \cite{liu2024}. SPARQL-LLM generates queries over federated life science endpoints using endpoint metadata and checks them against schema information before execution \cite{smeros2026}.

Beyond query generation, several methods explicitly check or constrain agent behavior. The Ontology-based Query Check flags queries that conflict with an ontology and prompts LLMs to revise them, which is reported to increase accuracy on an enterprise question answering benchmark \cite{allemang2024}. NaVQA addresses queries that use ontology terms without mappings to the underlying data, which can silently produce empty results, by restricting the agent to ontology terms represented in the data \cite{xiao2026}. These methods differ in what they check and how they intervene, but are typically evaluated by overall response accuracy. We instead vary the source of the ontology and evaluate each flag against the correct answer, distinguishing detected errors from false alarms.

Ontology axioms can be used to validate data, by interpreting them as constraints rather than only as a means for inference. Motik et al.\ showed that OWL axioms that resemble integrity constraints do not act as checks under standard first-order semantics, motivating a separate formalization for integrity constraints \cite{motik2009}. The W3C subsequently standardized SHACL for validating RDF data against structural constraints \cite{knublauch2017}, and ShEx offers a comparable language \cite{prudhommeaux2014}. Kontokostas
et al.\ translated vocabulary axioms such as domain, range, and disjointness into SPARQL tests over Linked Data, and found that domain and range tests identified the most errors \cite{kontokostas2014}. At the query level, Seifer et al.\
type-check SPARQL queries against an ontology and reject queries whose results
must be empty on any database \cite{seifer2019}.

However, ontology-based checks depend on the suitability of the ontology itself. Linked Data uses terms without defined vocabularies \cite{hogan2010}, while combining widely used vocabularies under a closed-world interpretation can expose conflicting definitions \cite{feeney2018}. In the biological domain, ontology-based constraints were used to evaluate annotations, which revealed limitations in the ontology itself \cite{tawfiq2026}. We ask the same question of checks on agents' answers at query time, where an unsuitable ontology flags correct answers.

Ontologies and constraints may be generated by LLMs rather than domain experts. LLMs generated shape schemas for Wikidata and YAGO which were evaluated against reference schemas \cite{zhang2025}. Shapes mined from large knowledge graphs were filtered according to how frequently the data supports them \cite{rabbani2023}, and constraints extracted from ontologies, mappings, schemas, and data were combined and used to evaluate a dataset \cite{duan2024}. In each case, the constraints are assessed against a reference or the data, not by how they perform as checks on downstream tasks.

Checks are useful only when their feedback is reliable. LLM self-correction is more effective when guided by reliable external feedback than by the model's own judgement, except in tasks particularly suited to self-evaluation \cite{kamoi2024}. Consequences of unreliable checks have also been studied directly, where ErrorLLM measures how often refining a correct SQL query flagged by a detector changes its result \cite{hong2026}. In software engineering, static analyses used in Google's code review are required to keep effective false positives below 10\%, reflecting the risk that false alarms erode developers' trust \cite{sadowski2018}. Queries that remain valid while expressing a different question than intended have been catalogued for SQL \cite{brass2006}, and LLMs detect them in generated SQL with limited accuracy \cite{liu2025}.

Evaluating checks requires care in how test cases are constructed. Artificially introduced faults provide controlled test cases, but may not represent the full range of real faults \cite{just2014}, and conclusions drawn from artificial faults do not always generalize to real ones \cite{pearson2017}. Evaluation can also be affected by information leakage, where models may recall identifiers from public knowledge graphs even when those identifiers are masked in the input \cite{gashkov2025}. Building on these lessons from software testing, SQL, and code, we evaluate deterministic, ontology-derived checks for SPARQL agents on both real agent answers and artificial faults.

%% file: sections/checks.tex
\section{Ontologies as Checks}
\label{sec:checks}

\subsection{Ontology sources}
\label{sec:sources}

We compare three types of ontologies: A \emph{curated} ontology is published by the knowledge graph's authors, and we use it exactly as published to reflect what an agent developer would obtain in practice. A \emph{data-derived} ontology summarizes the data itself, including the number of instances of each class, and, for each property, the classes of the instances that appear as its subjects and objects. Instances are counted under each of their asserted classes and their superclasses. A pattern is considered allowed only if it occurs at least five times in the data. It states domain and range but no disjointness axioms. An \emph{LLM-generated} ontology is produced by Claude Opus 5.5 from the same summary used to construct the data-derived ontology, supplemented with a fixed sample of instances per class. Thus, the difference between
these data-derived and LLM-generated ontologies is what the LLM adds, given the summary and a sample of
instances. We also evaluate a baseline with no ontology.

\subsection{Ontology uses}
\label{sec:uses}

Each ontology is used in three ways. \emph{As context}, the ontology is rendered as text in
the agent's prompt. All ontology sources use the same rendering procedure where classes and hierarchy are provided first, followed by domain and range, disjointness, and labels, all within a common token budget. \emph{As a check} on the agent's answer, each ontology is compiled using the same compiler, so differences between sources do not reflect differences in implementation. \emph{Both}: the same ontology is supplied as context in the prompt and used as a check on the answer.

\subsection{Check types}
\label{sec:check-types}

Checks examine either the query or its bindings, the solution mappings of its graph pattern. Query checks parse the SPARQL query and infer, from its triple patterns, the types required of each variable and constant, which they test against the ontology. The subject of a property must be an instance of the property's declared domain or one of its subclasses, its object must be an instance of the declared range or one of its subclasses, and no variable may be required to belong to two classes declared disjoint. A heuristic check, on in every condition, also flags a variable required to belong to two classes that share no subclass in the ontology, a closed-world counterpart of a missing disjointness axiom. A separate \revg{declaration} check requires every class and property named by the query to be declared in the ontology. \revg{A datatype check flags a literal outside the datatype a property declares as its range. The data-derived ontology adds pattern checks on co-occurrence: they flag a property used with a class the data never combines with it, and a join of two triple patterns that no class supports at least five times.}

Binding checks re-evaluate the query's graph pattern with every variable projected, collect the resulting bindings, and test the asserted types of the bound resources against the same axioms. Collection is limited to 10,000 triples per answer.

\subsection{Axiom interpretation}
\label{sec:axiom-interpretation}
We interpret ontology axioms as constraints on the data as it stands, following the closed-world reading of OWL axioms as integrity constraints \cite{motik2009}. A flag therefore indicates that the asserted data violates an axiom as written. Under OWL's open-world semantics \cite{owl2012}, the same data might instead entail a class membership that has not been explicitly asserted. We do not resolve this difference between interpretations, and instead quantify it by attributing each false alarm to its cause (RQ2, Section~\ref{sec:results}).

We distinguish \emph{contradictions} from \emph{mismatches}. A contradiction is a violation under both readings, where the data witnesses an explicit logical clash, such as an instance belonging to two disjoint classes, a value violating a declared datatype, or a functional property with two distinct literal values for the same subject. A mismatch is a violation under the closed-world reading only, such as a subject with no asserted type in a property's declared domain, which OWL would instead use to infer that type.

\subsection{Check verdicts}
\label{sec:outcomes}

A check \emph{passes} when the available evidence satisfies the constraint, and flags when it detects a violation. It returns \emph{unknown} when the evidence needed to decide is unavailable, for example because the value is untyped, retrieval reaches its limit, or the query contains a construct outside the supported SPARQL fragment (Appendix~\ref{sec:app-checks}). A check is \emph{not applicable} when there is no relevant axiom to test against the terms used in the query.

An answer is flagged if any check flags it. If none flags the answer but
at least one returns unknown, the answer is classified as unknown; otherwise,
it is passed. Unknown answers are not flagged and therefore do not trigger
a retry. They count as unflagged in every rate, and their share is
reported separately.
 
\begin{figure}[t]
  \footnotesize
  \begin{tabular}{@{}p{0.96\columnwidth}@{}}
    \toprule
    \textbf{Curated ontology (excerpt)}\\
    \texttt{ex:birthDate rdfs:domain ex:Person .}\\
    \texttt{ex:Person owl:disjointWith ex:City .}\\
    \texttt{ex:birthPlace rdfs:domain ex:Person .}\\
    \texttt{ex:birthPlace rdfs:range ex:Place .}\\
    \texttt{ex:population} is used by the data but not declared.\\
    \textbf{Data (excerpt)}\\
    \texttt{ex:Dublin rdf:type ex:City .}\\
    \midrule
    \textbf{Caught.} \emph{``When was Dublin founded?''}\\
    \texttt{SELECT ?d WHERE \{ ex:Dublin ex:birthDate ?d \}}\\
    Dublin is a city, and a city cannot have a birth date: flag. Running
    the query also returns nothing.\\
    \midrule
    \textbf{False alarm.} \emph{``What is the population of Dublin?''}\\
    \texttt{SELECT ?p WHERE \{ ex:Dublin ex:population ?p \}}\\
    The query is right, but the ontology does not declare
    \texttt{ex:population}: flag. A data-derived ontology passes it.\\
    \midrule
    \textbf{Missed.} \emph{``Who founded Dublin?''}\\
    \texttt{SELECT ?p WHERE \{ ?p ex:birthPlace ex:Dublin \}}\\
    The query returns the people born in Dublin. Every type fits the
    ontology, so every check passes a wrong answer.\\
    \bottomrule
  \end{tabular}
  \caption{Three answers checked against one curated ontology, in an invented vocabulary.}
  \label{fig:example}
  \Description{Three SPARQL queries over an invented vocabulary: one wrong
  query is caught by a domain and disjointness check, one correct query is
  flagged because a property is undeclared, and one wrong query passes
  every check.}
\end{figure}

\subsection{Scoring checks}
\label{sec:scoring}
 
There is a pre-defined correct answer for each question, and each check decision therefore
falls into one of four categories. A \emph{caught} answer is a flagged wrong answer. A \emph{missed} answer is an unflagged wrong answer. A \emph{false alarm} (a false positive) is a flagged correct answer. A \emph{correctly passed} answer is an unflagged correct answer. Figure~\ref{fig:example} shows a caught answer, a false alarm and a missed answer. We report the \emph{catch rate} (recall) as the proportion of wrong answers that are flagged, and the \emph{false-alarm rate} as the proportion of
correct answers that are flagged.
 
\subsection{Retries}
\label{sec:retry}
 
A flag triggers one retry. The agent resumes its conversation, and a fixed instruction tells it that its earlier answer has been set aside and asks for a final answer. It receives one of two messages: a \emph{specific} message, which identifies the failed constraint and the part of the query that violated it without revealing the ontology source, or a \emph{generic} message, ``Please re-check your answer.'' The two messages are alternative conditions applied from the same state, so the message is the only difference between them. The retry has the same tools and limits as the first attempt, and its answer is final. A retry that fails or abstains counts as wrong.
 
We classify a wrong answer that becomes correct after retry as \emph{fixed}, and a correct answer that becomes wrong as \emph{corrupted}. These outcomes measure both the benefit of a flag and the risk of acting on a false alarm.

%% file: sections/design.tex
\section{Study Design}
\label{sec:design}

Every confirmatory claim concerns one public graph, UniProt, and was registered with its test and margin before any test answer was examined (Section~\ref{sec:claims}). The other graphs, the human-written questions and every analysis not registered in advance are exploratory.

\subsection{Graphs}
\label{sec:graphs}

Table~\ref{tab:graphs} lists the five graphs used in the study. We use the Swiss-Prot subset from UniProt \cite{uniprot2025} r2026\_03 for all confirmatory claims. The curated ontology of this graph declares domain, range, and disjointness axioms, which allows all three kinds of checks to be evaluated. We use Rhea \cite{bansal2022}, which declares a class hierarchy but no domain, range or disjointness axioms in its curated ontology, for class membership checks. We also use a \revg{masked} copy of Rhea, in which we replace class, property, and instance identifiers and labels with opaque identifiers, to test whether agents and LLM-generated ontologies rely on recognising a familiar vocabulary. We also use a corporate graph from outside the life sciences
\cite{text2sparql} and a private graph of nearly $10^9$ triples with its own
curated ontology. Each graph is served from a triple store that
answers queries over its asserted triples without reasoning, and the public
graphs from QLever \cite{bast2017}. Results on graphs other than UniProt are exploratory, and we only report rates on
the private graph.

\begin{table}[t]
  \caption{The graphs: size, what the curated ontology declares, and role.}
  \label{tab:graphs}
  \footnotesize
  \setlength{\tabcolsep}{2.5pt}
  \renewcommand{\arraystretch}{0.9}
  \begin{tabularx}{\linewidth}{@{}lrXl@{}}
    \toprule
    Graph & Triples & Curated ontology declares & Role \\
    \midrule
    UniProt & $4.0\times10^{8}$ & domain, range, disjointness & confirmatory \\
    Rhea & $5.5\times10^{6}$ & class hierarchy only & exploratory \\
    \revg{Rhea, masked} & $5.5\times10^{6}$ & as Rhea & exploratory \\
    Corporate & $2.7\times10^{4}$ & domain, range; no disjointness & exploratory \\
    Private & ${\sim}10^{9}$ & not described & rates only \\
    \bottomrule
  \end{tabularx}
\end{table}

\subsection{Ontologies}
\label{sec:ontologies}
Each graph has the three ontology sources defined in Section~\ref{sec:sources}, and a baseline without an ontology. For UniProt, every ontology fits within the common token budget and is given to the agent in full. We draw three LLM-generated ontologies per graph\revg{; on the masked copy of Rhea they are drawn from its masked data summary}. On UniProt we draw three more \emph{masked} ontologies from a copy of the data summary in which identifiers are renamed and string values masked; they are used only as checks. C3 uses these masked draws, so that agreement with the curated ontology does not merely reflect recall of familiar identifiers; masking reduces the cues for recalling UniProt's published ontology, but it cannot establish that the model has never seen it \revg{(closed-book probes, Appendix~\ref{sec:app-recall})}. Their domain and range axioms are a subset of the data-derived ontology's, and the LLM adds disjointness, so C3 compares the curated axioms with data-derived domain and range plus LLM-added disjointness. The properties of these draws are described in Appendix~\ref{sec:app-ontology-draws}.

Each ontology source can be used as context, as a check, or in both roles (Section~\ref{sec:uses}). On UniProt, every model answers with no ontology, the curated or the data-derived ontology in its prompt, and Sonnet and Qwen3.8 also with the first LLM-generated one. The checks of every source, including the masked ontologies and the combined check, are applied to the first answers of every prompt condition. Retries are run on the answers given with no ontology: after a curated flag, once with the specific message and once, from the same state, with a generic request to re-check, and after each comparator's flags (Section~\ref{sec:baselines}).

\subsection{Questions and correct answers}
\label{sec:questions}

Questions are generated from each graph's structure using seven templates, ranging from single lookups to compositional questions involving joins, filters and, aggregation. Each question is paired with the SPARQL query that answers it on the graph. The result of that query is the question's pre-defined correct answer. Agent answers are scored mechanically by exact match using scorer rules fixed in advance. A guard rejects ambiguous questions: for each template, plausible alternative readings, such as including narrower classes or transitive membership, are run against the graph, and a question whose answer changes under one must state its scope or is not generated. Each correct query is re-executed against the graph. An independent LLM agent, given only the question text and not its paired query, produced queries matching the correct answers for all 30 questions in a sample from an earlier version of the set.

Because agents answered simple lookups almost perfectly on development data, the compositional tier was made harder in at most two rounds, each raising every compositional template one step along one axis, aiming at an accuracy of about 50--80\% for Sonnet without an ontology on the development questions. For each graph, questions are divided into development (30\%) and test (70\%) sets within each template, resulting in 210 test questions on UniProt, 70 lookups and 140 compositional. In addition, we use 173 UniProt \revg{and 124 Rhea} question--query pairs from a published collection~\cite{bolleman2025} as a reference set that models may have seen during training\revg{, run for two models (Appendix~\ref{sec:app-human})}.

\subsection{Agents}
\label{sec:agents}

Seven models answer the questions: two proprietary models, Claude Sonnet~5.5
and Claude Haiku~5.5, and five open-weight models, Qwen3.8-27B,
Qwen3-Coder-30B-A3B, GLM-5.3-Flash, Gemma-4-31B and DeepSeek-V4-Flash. Each
agent queries the graph through a SPARQL tool and submits its answer as rows of
identifiers and values. Each question is attempted three times per model and
condition, as independent runs, subject to turn and time limits held constant across models.

\subsection{Artificial faults}
\label{sec:bank}

Agent errors are sparse and unevenly distributed, so we also introduce artificial faults in correct queries to measure checks independently of agent behaviour. Each fault applies a single operator to the correct query of a UniProt test question or of an additional question generated with the same templates, resulting in 1,260 questions. The operators form three classes: (1) invented identifiers absent from the graph. (2) swaps of a class, property, or constant for an item of a different type\revg{, drawn from the data-derived ontology's co-occurrence graph, whose nodes are properties and the classes and datatypes the data uses with them at least five times, and whose edges join a property to such a class or datatype. A swap replaces a class or property with a node of the same kind exactly two or four steps away, so that at two steps two classes share a property or two properties share a class; a constant is replaced by a member of a class at that distance from its own classes}. (3) meaning changes that preserve the query form, such as replacing \texttt{COUNT} with \texttt{COUNT DISTINCT}, \texttt{MIN} with \texttt{MAX}, or dropping a grouping key. Swaps of a constant for another constant of the same type are reported as a separate stratum, which accounts for the four rows in Table~\ref{tab:faults}.

Each faulty query is labelled according to whether it changes the answer. Unmodified correct queries and rewrites that preserve the answer are controls used to count false alarms. To evaluate any check without running an agent, the check is applied to each faulty query and each control. A flag on a faulty query counts as a catch only if the fault changes the answer, and a flag on a control counts as a false alarm. Since a query that returns no rows already signals a problem, rates are also reported separately for faults whose queries still return rows, and compared with retrying only when a query returns no rows. Table~\ref{tab:faults} reports the composition of the fault set.

\subsection{Claims and statistics}
\label{sec:claims}
The claims, outcomes, tests, and margins were registered and frozen in a timestamped commit before any test question was run. Each claim answers one of the research questions, as labelled below.
\begin{description}[leftmargin=0em,topsep=1pt,partopsep=0pt]
  \item[RQ1, C2] As checks, compare the curated and data-derived ontologies on (i) false-alarm rate for correct queries and (ii) catch rate for type swap faults. C2 requires both differences in the same direction, each with a five-percentage-point margin.
  \item[RQ1, C3] As checks, compare the curated ontology with the mean over the three masked LLM-generated ontologies on false-alarm rate, with a five-percentage-point margin. This is the central comparison.
  \item[RQ3, C1] As context, compare curated and data-derived ontologies on agents' first-answer accuracy: C1a tests for a difference\revg{, pooled over the seven models,} and C1b tests equivalence within $\pm3$ accuracy points \revg{on Qwen3.8 and Sonnet, the two models fixed for it in advance, because a pooled equivalence could hide opposite effects in different models}.
  \item[RQ3, C4] Compare the combined curated and data-derived checks with each check alone on net detection gain when wrong answers make up 25\% of answers.
  \item[RQ3, C5] After a flag, compare the check's specific message with a generic request to re-check on accuracy after one retry.
\end{description}
C1a, C2 and C3 are two-sided tests, and C4 and C5 test for a gain. The margins are the smallest differences of interest; for C1b, and for C3 when read from its interval, they are equivalence bounds.

We control the family-wise error rate at $\alpha = 0.05$ with a weighted Holm procedure \cite{holm1979}, with
the most weight on C3, fixed in the pre-registration from simulations without
any test data and after an independent statistical review. A claim is significant by the weighted-Holm
decision on its calibrated $p$-value, written $p^*$; the unadjusted bootstrap
$p$-values are in Appendix~\ref{sec:app-sensitivity}. Non-significance establishes neither equivalence
nor the absence of a benefit. Each test is a paired cluster
bootstrap whose nominal level was calibrated on simulated data before the
freeze, with the question as the unit of analysis. From the same simulations,
C1b, C4 and C5 were declared in advance as expected to be inconclusive at
their margins; we report them as bounds. Every other analysis is exploratory
and labelled as such. Inference is conditional on the seven frozen question templates, the
frozen artificial faults and the recipe for generating ontologies. Changes made after the freeze preserve comparisons within each model and graph and leave every registered decision unchanged; \revg{the Ontology-based Query Check, registered also on the artificial faults, was run on real answers only.}

\subsection{Baselines}
\label{sec:baselines}

In addition to retrying when a query returns no rows, four comparators run in the same harness: (1) an LLM judge given the question, query, results, and answer, implemented by the agent's own model for agent answers and by Claude Sonnet~5.5 and Haiku~5.5 for artificial faults; (2) SPARQL-LLM query validation~\cite{smeros2026}, as published; (3) the Ontology-based Query Check rules~\cite{allemang2024}, run from its paper's printed rules; and (4) trivial checks that always or never flag. Results are reported in the appendix.

%% file: sections/results.tex
\section{Results}
\label{sec:results}

We address the three research questions in turn, using the UniProt knowledge graph unless stated otherwise. The real answer analysis includes the first answers to 210 test questions from seven LLM models, with three attempts \revg{per question}, involving 4,410 answers. Of these, 3,383 contained a query that could be checked. We calculate exploratory rates per answer, pooled \revg{across} models. For \revg{artificial} faults, we pool rates across 9,753 \revg{artificial} faults that change the answer, and 3,630 correct \revg{queries}. Confirmatory comparisons weigh questions equally with 95\% confidence intervals from a paired cluster bootstrap over questions.

\subsection*{RQ1: \revg{Curated checks raise more false alarms than data-derived and masked LLM-generated checks}}
\label{sec:rq1}

\revg{RQ1 examines which errors each type of ontology catches when used as a check, how often it flags correct answers, and at what cost. The curated ontology raises more false alarms than both the data-derived ontology and the LLM-generated ontologies built from a summary with masked identifiers. On real agent answers, its flags are only marginally more precise than random flags.} Table~\ref{tab:main} reports catch and false alarm rates for each ontology on real agent answers and \revg{artificial} faults. The first row shows a baseline that retries any query that returns no rows and does not require an ontology. Ontology-based checks only add value where they catch errors beyond this baseline. \revg{All checks read about the same number of triples per answer, so their costs mainly differ due to retries after a flag. The curated ontology check triggers more retries than the data-derived ontology check, incurring greater token usage and runtime (Appendix~\ref{sec:app-cost}).}

Exploratory results on real answers reveal trade offs between catching errors and flagging correct answers. The curated ontology catches more wrong answers than the data-derived and masked LLM-generated ontologies, but most of its flags concern correct answers, and its flags are right hardly more often than those of a random flag at the same rate (Table~\ref{tab:comparisons}). The data-derived and masked ontologies flag fewer correct answers and are more likely to be right when they flag an answer, but each catches fewer than one in ten wrong answers. The three LLM-generated ontologies written from the original summary vary from one generation to the next: two catch few wrong answers, whereas the third catches more than the curated ontology but also flags far more correct answers (Table~\ref{tab:main}).

All ontologies catch a smaller proportion of errors in real answers than in \revg{artificial} faults. Most \revg{artificial} faults produce empty query results, whereas almost all wrong agent answers \revg{return} plausible results. Consequently, the empty result baseline catches \revg{few} real wrong answers. Retrying flagged answers \revg{with the} curated check's specific feedback corrects \revg{more than a quarter of the flagged wrong answers but corrupts about one in a hundred flagged correct answers (Table~\ref{tab:main})}.

\begin{table*}[t]
  \caption{Each ontology as a check on UniProt, in \%: real answers of seven models, and artificial faults. Bold, per column: the highest catch rate, precision and share fixed; the lowest false-alarm rate, share unknown and share corrupted.}
  \label{tab:main}
  \footnotesize
  \setlength{\tabcolsep}{2.5pt}
  \renewcommand{\arraystretch}{0.9}
  \begin{tabular*}{\linewidth}{@{\extracolsep{\fill}}lcccccccc@{}}
    \toprule
    & \multicolumn{3}{c}{Real answers} & \multicolumn{4}{c}{Artificial faults}
    & \\
    \cmidrule(lr){2-4}\cmidrule(lr){5-8}
    Check from & catch & false alarm & unknown & catch, all & catch, rows
    returned & false alarm & precision at 25\% wrong & fixed / corrupted \\
    \midrule
    Retry on empty result (reference) & 2.5 & \textbf{0.1} & --- & 79.2 & --- & \textbf{0.0} & \textbf{100} & 15.4 / \textbf{0.0} \\
    Curated ontology, specific message & 18.9 & 16.8 & \resLedgerUniprotNoneCheckerCuratedUnknownRate & \resLedgerUniprotNoneCheckerCuratedBankRecall & \resLedgerUniprotNoneCheckerCuratedBankNonEmptyRecall & \resLedgerUniprotNoneCheckerCuratedBankFalseFlagRate & \resUniprotBankBankReplayCuratedPrecisionAtPrevalenceTwoFive & 28.3 / 1.0 \\
    \quad same flags, generic message & \revg{18.9} & \revg{16.8} & \revg{\resLedgerUniprotNoneCheckerGenericUnknownRate} & \revg{\resLedgerUniprotNoneCheckerCuratedBankRecall} & \revg{\resLedgerUniprotNoneCheckerCuratedBankNonEmptyRecall} & \revg{\resLedgerUniprotNoneCheckerCuratedBankFalseFlagRate} & \revg{\resUniprotBankBankReplayCuratedPrecisionAtPrevalenceTwoFive} & 20.2 / 0.2 \\
    \quad class checks only & 3.1 & 0.6 & --- & 17.0 & --- & 1.1 & 83.4 & --- \\
    \quad class, domain and range & 8.0 & 8.8 & --- & 26.8 & --- & 6.1 & 59.5 & --- \\
    Data-derived ontology & 8.2 & 3.5 & \resLedgerUniprotNoneReplayCensusUnknownRate & \resLedgerUniprotNoneReplayCensusBankRecall & \resLedgerUniprotNoneReplayCensusBankNonEmptyRecall & \resLedgerUniprotNoneReplayCensusBankFalseFlagRate & \resUniprotBankBankReplayCensusPrecisionAtPrevalenceTwoFive & \revg{7.0 / 1.0} \\
    LLM-generated, draw 1 & 6.7 & 3.2 & \resLedgerUniprotNoneReplayGeneratedOneUnknownRate & 54.5 & 46.7 & 1.2 & 93.7 & \revg{11.4 / 24.2} \\
    LLM-generated, draw 2 & 8.6 & 6.6 & \resLedgerUniprotNoneReplayGeneratedTwoUnknownRate & 57.1 & 47.5 & 6.3 & 75.3 & \revg{13.3 / 12.7} \\
    LLM-generated, draw 3 & \textbf{43.4} & 24.3 & \resLedgerUniprotNoneReplayGeneratedThreeUnknownRate & 69.6 & 53.7 & 29.4 & 44.1 & \revg{\textbf{30.8} / 6.0} \\
    LLM-generated, masked (three) & 5.0 & 1.8 & \resLedgerUniprotNoneReplayGeneratedDcOneUnknownRate & \resLedgerUniprotNoneReplayGeneratedDcOneBankRecall--\resLedgerUniprotNoneReplayGeneratedDcTwoBankRecall & \resLedgerUniprotNoneReplayGeneratedDcOneBankNonEmptyRecall--\resLedgerUniprotNoneReplayGeneratedDcTwoBankNonEmptyRecall & \resLedgerUniprotNoneReplayGeneratedDcOneBankFalseFlagRate--\resLedgerUniprotNoneReplayGeneratedDcTwoBankFalseFlagRate & \resUniprotBankBankReplayGeneratedDcTwoPrecisionAtPrevalenceTwoFive--\resUniprotBankBankReplayGeneratedDcOnePrecisionAtPrevalenceTwoFive & not run \\
    Curated and data-derived combined & 19.5 & 17.8 & \textbf{\resLedgerUniprotNoneReplayUnionUnknownRate} & \textbf{80.0} & \textbf{67.3} & 23.6 & 53.1 & \revg{29.4 / 1.0} \\
    \bottomrule
  \end{tabular*}
\end{table*}

\revg{On the artificial faults, no single ontology catches as many faults as the retry on empty results, which raises no false alarm (Table~\ref{tab:main}).} The two confirmatory
comparisons for RQ1 are decided on the artificial faults, per question (Table~\ref{tab:comparisons}). As checks, the data-derived ontology raises fewer false alarms on the control queries than the curated ontology and catches more type swaps, both by more than the registered 5-point margin (C2). The curated ontology also raises more false alarms than the mean of the masked LLM-generated ontologies, by more than the margin (C3).

\begin{table}[t]
  \caption{Registered claims and exploratory comparisons on UniProt, in points, with 95\% intervals; for the claims, the margin and the adjusted $p$ ($p^*$). Bold: established.}
  \label{tab:comparisons}
  \scriptsize
  \setlength{\tabcolsep}{2.5pt}
  \renewcommand{\arraystretch}{0.9}
  \begin{tabularx}{\linewidth}{@{}Xccc@{}}
    \toprule
    Comparison & Estimate (95\% CI) & Margin & $p^*$ \\
    \midrule
    \multicolumn{4}{@{}l}{\emph{RQ1: as checks, on artificial faults}} \\
    C2: false alarms on controls, curated $-$ data-derived & \textbf{\boldmath $+8.6$ (6.0, 11.2)} & 5 & \resConfCTwoP \\
    \quad type-swap catches, curated $-$ data-derived & \textbf{\boldmath \resConfCTwoTypeWrongRecallCuratedCensusEst{} (\resConfCTwoTypeWrongRecallCuratedCensusLo, \resConfCTwoTypeWrongRecallCuratedCensusHi)} & 5 & \\
    C3: false alarms on controls, curated $-$ masked LLM-generated & \textbf{\boldmath \resConfCThreeCuratedMeanMaskedEst{} (\resConfCThreeCuratedMeanMaskedLo, \resConfCThreeCuratedMeanMaskedHi)} & 5 & \resConfCThreeP \\
    \multicolumn{4}{@{}l}{\emph{RQ3: accuracy, and net detection for C4}} \\
    C1a: context, curated $-$ data-derived & \textbf{\boldmath \resConfCOneAPooledOverPEst{} (\resConfCOneAPooledOverPLo, \resConfCOneAPooledOverPHi)} & --- & \resConfCOneAP \\
    C1b: the same, Qwen3.8 & \resConfCOneBQwenEst{} (\resConfCOneBQwenLo, \resConfCOneBQwenHi) & $\pm3$ & \resConfCOneBP \\
    \quad Sonnet & \resConfCOneBSonnetEst{} (\resConfCOneBSonnetLo, \resConfCOneBSonnetHi) & $\pm3$ & \\
    C4: combined $-$ curated check & \resConfCFourUnionCuratedEst{} (\resConfCFourUnionCuratedLo, \resConfCFourUnionCuratedHi) & 1 & \resConfCFourP \\
    \quad combined $-$ data-derived check & \resConfCFourUnionCensusEst{} (\resConfCFourUnionCensusLo, \resConfCFourUnionCensusHi) & 1 & \\
    C5: specific $-$ generic message, per flagged answer & \resConfCFivePooledOverPEst{} (\resConfCFivePooledOverPLo, \resConfCFivePooledOverPHi) & 10 & \resConfCFiveP \\
    \midrule
    \multicolumn{4}{@{}l}{\emph{Exploratory, no test}} \\
    C2's type swaps two steps apart & $-1.3$ ($-2.6$, 0.1) & & \\
    C5 weighted by each model's flags & $+0.7$ ($-1.2$, 3.0) & & \\
    Real answers: precision $-$ random flag, curated & $+1.6$ ($-2.3$, 6.3) & & \\
    \quad the same, LLM judge & $+15.8$ (12.5, 19.3) & & \\
    Curated check: curated $-$ data-derived in the prompt & $-14.3$ ($-19.9$, $-9.3$) & & \\
    Data-derived check: data-derived $-$ curated in the prompt & $-4.8$ ($-8.7$, $-1.0$) & & \\
    \bottomrule
  \end{tabularx}
\end{table}

Two features of the design favour the data-derived ontology in C2. It is built from the graph on which the checks run, so it cannot flag a pattern that occurs at least five times in the data. The type swaps were also generated from its co-occurrence graph, so nearly all of them form patterns absent from the data, which executing the query would also reveal (Table~\ref{tab:main}). The
difference in type swap catch rate arises from swaps four steps apart on the
co-occurrence graph; for swaps two steps apart it is not established (exploratory; Table~\ref{tab:comparisons}). A lower false-alarm rate does
not by itself make a better check: the \revg{masked} ontologies also catch
fewer of the faults whose queries still return rows than either the curated or
the data-derived ontology.

\revg{Sonnet and Qwen3.8 are far less accurate on the human-written questions than on the generated ones, so these leave checks much more to catch. On the UniProt questions Sonnet's curated check beats a random flag; on the Rhea questions it does no better than one (Appendix~\ref{sec:app-human}).}
Among the comparators (Appendix~\ref{sec:app-baselines}), an LLM judge, the agent's own model reading the question, the query, the results and the answer, catches more wrong answers than either the curated or the data-derived ontology at a false-alarm rate similar to the curated check's, and its flags are right more often than a random flag's (Table~\ref{tab:comparisons}). Acting on its flags was not beneficial: retries informed by the judge's reason corrupted more flagged correct answers than they fixed wrong ones, unlike the curated check's.
Results per model and on the other graphs are in Appendices~\ref{sec:app-models} and~\ref{sec:app-graphs}.

Table~\ref{tab:faults} shows the distribution of artificial faults by class. The classes
differ in how often the faulty query still returns rows, \revg{and so in how much a check can add to running the query}. Invented identifiers and type swaps
mostly do not return rows, so a retry on empty results already catches most of
them\revg{; the ontologies differ mainly on type swaps}. Constants swapped for others of
the same type and changes of meaning mostly still return rows, and every
ontology catches far fewer of them than a random flag at the curated check's
rate would.

\begin{table}[t]
  \caption{Catch rates by class of artificial fault on UniProt, in \%. LLM-m: masked LLM-generated, mean of three; rand.: a random flag at the curated check's rate. Bold: a random flag catches more than every ontology.}
  \label{tab:faults}
  \footnotesize
  \setlength{\tabcolsep}{2.5pt}
  \renewcommand{\arraystretch}{0.9}
  \begin{tabular*}{\linewidth}{@{\extracolsep{\fill}}lrcccccc@{}}
    \toprule
    Class & $n$ & no rows & cur. & data & LLM-m & rand. \\
    \midrule
    Invented identifier & 2,986 & 80.5 & 77.4 & 77.5 & 74.5 & 48.1 \\
    Type swap & 6,413 & 80.0 & 55.3 & 78.4 & 49.9 & 48.0 \\
    Same-type constant & 1,554 & 46.8 & 3.1 & 5.8 & 0.0 & \textbf{47.9} \\
    Change of meaning & 245 & 41.2 & 0.0 & 0.0 & 0.0 & \textbf{44.9} \\
    \midrule
    Correct items (false alarm) & 3,630 & 0.0 & \resLedgerUniprotNoneCheckerCuratedBankFalseFlagRate & \resLedgerUniprotNoneReplayCensusBankFalseFlagRate & \resLedgerUniprotNoneReplayGeneratedDcOneBankFalseFlagRate--\resLedgerUniprotNoneReplayGeneratedDcTwoBankFalseFlagRate & --- \\
    \bottomrule
  \end{tabular*}
\end{table}

\subsection*{RQ2: \revg{Domain, range and declaration checks raise the curated ontology's false alarms, and ontology checks miss changes of meaning}}
\label{sec:rq2}

\revg{RQ2 asks what causes the false alarms and which errors the checks miss. The curated ontology's false alarms come from its domain and range axioms and its requirement that every term be declared, and each curated ontology we studied disagrees with its own data; the curated, data-derived and masked ontologies catch none of the changes of meaning among the artificial faults.} Evaluating the curated ontology one axiom type at a time (Table~\ref{tab:main}, the curated rows) shows that class membership and subsumption alone produce almost no false alarms; adding \texttt{rdfs:domain} and \texttt{rdfs:range} axioms raises them, and requiring every class and property in the query to be declared raises them the most. On real answers, domain and range checks flag correct and incorrect answers at similar rates, so their flags are right no more often than random ones, while the flags from class membership and term declarations do better than random (Appendix~\ref{sec:app-layers}). In the curated and masked ontologies, disjointness axioms do not contribute to flags: neither the seven pairs of the curated ontology nor the 73 to 759 pairs added by the LLM in the masked ontologies flag any answer not already flagged by another axiom type. C3's difference therefore rests on domain and range axioms and term declarations. The data-derived ontology's checks are analysed in the same way in Appendix~\ref{sec:app-layers}.

Table~\ref{tab:schema-data} shows where each curated ontology disagrees with its own data: with logical contradictions in UniProt, with omissions of properties the data uses extensively and the graph's documentation describes in Rhea, and mostly through naming on the private graph, whose curated checks flag most correct answers (Appendix~\ref{sec:app-graphs}).

\begin{table}[t]
  \caption{Where a curated ontology, as published, disagrees with its own data.}
  \label{tab:schema-data}
  \footnotesize
  \setlength{\tabcolsep}{2.5pt}
  \renewcommand{\arraystretch}{0.9}
  \begin{tabularx}{\linewidth}{@{}lX@{}}
    \toprule
    Graph & Disagreement \\
    \midrule
    UniProt & Contradictions: \texttt{up:maximum} is declared \texttt{xsd:float},
      but all 139 of its values are integers; 5 \revg{data} properties declared functional
      have two \revg{distinct} values for some subjects (82 entries for one property, 3,859
      citations for four others). No instance of two disjoint classes. \\
    Rhea & Omissions: 4 properties the data uses (between 15,032 and 90,295
      times each) are not declared. \\
    Private & 94.2\% of triples use an identifier other than the ontology's
      own, mostly renamings of declared terms; 19.9\% use a term with no
      counterpart in it; 1.46\% have a subject typed outside the property's
      declared domain (a mismatch). \\
    \bottomrule
  \end{tabularx}
\end{table}

\revg{In a blind audit of a sample of the curated check's false alarms, two annotators judged the cause of each (Table~\ref{tab:why}, top; Appendix~\ref{sec:app-audit}).} Most came from a term the ontology does not declare and most of the rest from a closed-world reading, and every audited false alarm was indeed a correct answer.

On the artificial faults, the curated, data-derived and
masked ontologies catch none of the changes of meaning in Table~\ref{tab:faults} and few of the constants swapped for others of the same type, whereas a random flag at the curated check's rate would catch many of both. One LLM-generated ontology built from the
original summary catches 25.7\% of the changes of meaning and 17.2\% of the
same-type swaps, but it also flags the most correct items (Table~\ref{tab:main}). An LLM judge, given the question, catches all 41 changes of meaning it judged (Appendix~\ref{sec:app-baselines}). On real answers, two annotators blind to the
model, the condition and the checks' verdicts, coded a sample of the wrong
answers that no check flagged (Table~\ref{tab:why}, bottom).
Most missed wrong answers took a wrong path through the graph or a wrong filter or scope, and almost none could be flagged by a structural check even over a correct schema.

\begin{table}[t]
  \caption{Why checks go wrong, \revg{in a blind audit of a sample: the curated check's false alarms by judged cause (top) and missed wrong answers by kind of error (bottom), counted where both annotators gave the same code, out of 32 and 64 items; Cohen's $\kappa$ 0.77 and 0.54.}}
  \label{tab:why}
  \footnotesize
  \setlength{\tabcolsep}{2.5pt}
  \renewcommand{\arraystretch}{0.9}
  \begin{tabularx}{\linewidth}{@{}l*{6}{>{\centering\arraybackslash}X}@{}}
    \toprule
    \revg{False alarm} & \revg{undeclared term} & \revg{closed world} & \revg{data clash} & \revg{schema too narrow} & \revg{checker defect} & other \\
    \midrule
    \revg{Curated} & 17 & 11 & 0 & 0 & 0 & 0 \\
    \bottomrule
  \end{tabularx}

  \smallskip
  \setlength{\tabcolsep}{1.5pt}
  \begin{tabularx}{\linewidth}{@{}l*{8}{>{\centering\arraybackslash}X}@{}}
    \toprule
    Missed & aggre\-gation & wrong constant & filter or scope & path & shape & \revg{gold or question} & \revg{scorer} & other \\
    \midrule
    Answers & 0 & 2 & 13 & 21 & 3 & 1 & 1 & 0 \\
    \bottomrule
  \end{tabularx}
\end{table}

\subsection*{RQ3: \revg{Data-derived context does better; combining and specific messages show no gain}}
\label{sec:rq3}

RQ3 asks whether the ontology improves performance beyond its use as a single
check, and we examine three uses: providing the ontology as context to the agent, combining checks derived from two ontology sources, and including the check's specific message when the agent retries a flagged answer. As context, the data-derived ontology gives higher accuracy than the curated one; combining the two checks shows no net gain in detection, and passing the check's specific message to the agent no gain in accuracy, with intervals that bound any gain at a few points. Table~\ref{tab:comparisons} gives each estimate with its interval and registered margin.

As context, the data-derived ontology yields higher accuracy than the
curated ontology pooled over the seven models (C1a). Equivalence within 3 points is not established for Qwen3.8 and Sonnet, the two models fixed for it in advance (C1b).

A net detection gain from combining the curated and data-derived checks is not established (C4). The combination catches more faults than either check alone but also raises more false alarms than either (Table~\ref{tab:main}), so it does no better than the curated check alone and worse than the data-derived check alone.

The check's specific message does not
detectably improve on a generic request to re-check (C5), whether the seven models are weighted equally or by their flags. The specific message fixed more of the flagged wrong answers than the generic request, but also corrupted more of the flagged correct ones (Table~\ref{tab:main}). C1b, C4 and C5 were declared in advance as expected to be inconclusive, and we interpret them as bounds on the size of the effects, not as evidence that the effects are absent.

\revg{A setting an agent developer is likely to deploy uses} the data-derived ontology as context and the curated ontology as the check. On this configuration, the retry resulted in a small gain in accuracy, \revg{or none, for each of the seven models} (exploratory; Appendix~\ref{sec:app-graphs}). \revg{On the other graphs where retries were run the change was smaller or negative (Appendix~\ref{sec:app-graphs}).}

Using the same ontology as context and as the check is associated with fewer catches \revg{(exploratory)}. The curated check catches fewer of the wrong answers produced with the curated ontology in the prompt than of those produced with the data-derived ontology in the prompt; the data-derived check shows the same pattern with the roles reversed, more weakly (Table~\ref{tab:comparisons}).

%% file: sections/discussion.tex
\section{Discussion}
\label{sec:discussion}

\paragraph{Implications for knowledge graph publishers.}
Published, curated ontologies are not necessarily reliable as checks. On the published UniProt knowledge graph, ontologies derived from the data, either summarized directly or written by an LLM from masked identifiers at a cost of under one US dollar each (Appendix~\ref{sec:app-cost}), raised fewer false alarms than the curated ontology. These curated ontologies often disagree with the published data and knowledge graphs (Table~\ref{tab:schema-data}), and these disagreements have direct consequences. Because checks require every class and property used in a query to be declared, each property missing from the ontology produces a false alarm on every correct answer that uses it. An ontology is a reliable check only if it declares every term its data uses and its axioms are satisfied by that data. The curated ontologies used in this study do not meet this standard, and a summary derived from the data itself is sufficient to reveal where they fall short. Publishers could therefore release summaries alongside the ontology, or directly use the summary to validate the ontology before each release.

\paragraph{Implications for agent developers.}
For developers of agents, a check earns its place only on errors that running the query does not already
reveal, so any check should be compared with a retry on empty results before
it is deployed. False alarms deserve the same scrutiny as catches: static
analyses at Google are required to keep effective false positives below 10\% to retain
developers' trust \cite{sadowski2018}, and a check that sends an agent to
revise correct answers reduces trust in the same way. Measured comparably, as the proportion of a check's flags that fall on correct answers, every ontology check and the LLM judge exceed this threshold by a wide margin on the agents' answers to the generated questions, most of which are correct. On questions written by humans, where most answers are wrong, most flags fall on wrong answers. This proportion therefore depends on how often the agent errs, which is why we compare each check with a random check that flags the same proportion of answers. In the configuration a developer is most likely to deploy, with the data-derived ontology as context and the curated ontology as the check, retries improved accuracy slightly on UniProt, and yielded smaller gains or losses on the other graphs on which they were run (Section~\ref{sec:rq3}). An LLM judge detected more errors than either the curated or the data-derived ontology, but acting on its flags corrupted more correct answers than it fixed, and it requires a model call for every answer it checks, whereas an ontology-derived check requires none.

\paragraph{Limitations.}
The confirmatory claims apply for UniProt's frozen questions, artificial faults and
ontologies. The other graphs show how far they extend. \rev{The
LLM-generated ontologies were produced by a single model. Ontologies written by two open-weight
models flag substantially more correct queries on average and vary widely between draws (Appendix~\ref{sec:app-ontologies}).} On Rhea, whose curated ontology declares no domain, range or disjointness axioms, the curated check flags far more correct answers than on UniProt, whereas the data-derived ontology flags few (Appendix~\ref{sec:app-graphs}). We do not report results on Rhea's artificial faults, which were provisional and partly inspected during development. The difference between ontologies used as context that we
observed on UniProt was not replicated on Rhea. Finally, the private graph is
not publicly available, so its results cannot be independently reproduced, and
no conclusion rests on it alone.

%% file: sections/conclusion.tex
\section{Conclusion and Future Work}
\label{sec:conclusion}

An ontology used to check an LLM agent's answers is itself a judge that can
be wrong, and we measured how: on the same answers and the same artificial
faults, for a curated ontology, an LLM-generated one and one derived from the
data, with every comparison fixed before any test answer was seen. \revg{The curated ontology raised more false alarms than the data-derived one and the LLM-generated ones written from masked names, and on real answers its flags were right hardly more often than random ones.} \revg{Beyond checking, the data-derived ontology served the agents better as context, and neither combining the two checks nor passing the check's specific message to the agent showed a gain.}
Our artificial faults let anyone measure a new check's catches and false alarms without running an agent.

Our measurements rely on knowing the correct answer to every question, and real
use does not. We will apply these checks to agents that help biologists analyse
a large knowledge graph of genomes, proteins and the computational predictions
made about them, where no correct answer is known in advance and a flag is the
only automatic sign that an answer may be wrong. The rates measured here will decide which checks such an agent runs, beside a new check that every value traces back to its source.

%% file: sections/appendix.tex
\section{The replication package}
\label{sec:appendix}
\label{sec:app-package}

The replication package is available at \url{https://anonymous.4open.science/r/www-2027-replication-package-36D6}. \revg{It holds the code, under the MIT licence; the frozen inputs and configurations; each model's final settings; the commands that run the study; and the analysis that reproduces every number reported here, with the data under CC BY 4.0.} The private graph contributes rates only: its data, its alignment mapping, its generated ontologies and the checks' messages on it are not released.

\section{The method in detail}
\label{sec:app-method}

This part describes how the study was carried out, beyond Sections~\ref{sec:checks} and~\ref{sec:design}. It lists the checks, describes the LLM-generated ontologies, and gives the cost of checking.

\subsection{The checks in detail}
\label{sec:app-checks}

Table~\ref{tab:app-checks} lists every check, what it reads, and when it cannot decide. The checks are compiled from the ontology into SPARQL queries over the bindings of a query, and into tests on the parsed query. The supported fragment includes basic graph patterns, \texttt{OPTIONAL}, negation, aggregates, and fixed-length paths; constructs outside it, including \texttt{UNION} and subqueries, give an unknown verdict rather than a pass or a flag.  The numbers-traceable check needs no ontology and enters no comparison between ontologies.

\begin{table}[H]
  \caption{The checks: what each reads, and when it reports unknown.}
  \label{tab:app-checks}
  \scriptsize
  \setlength{\tabcolsep}{2.5pt}
  \renewcommand{\arraystretch}{0.9}
  \begin{tabularx}{\linewidth}{@{}XXX@{}}
    \toprule
    Check & Reads & Unknown when \\
    \midrule
    Term exists & the query & --- \\
    Class membership & query; bindings & a value has no type \\
    Domain and range & query; bindings & a value has no type \\
    Disjointness & query; bindings & a value has no type \\
    Datatype range & bindings & --- \\
    Numbers traceable & answer and results & --- \\
    \bottomrule
  \end{tabularx}
\end{table}

\subsection{The LLM-generated ontologies}
\label{sec:app-ontologies}

The three ontologies an LLM wrote from the masked UniProt summary\label{sec:app-ontology-draws} keep 87--90\% of the data-derived ontology's domain and range pairs and add none of their own, so their property sets almost coincide with it (Jaccard similarity 0.93), and they add 73, 759 and 77 disjointness pairs, for which the data-derived ontology has no layer. Their domain and range overlap little with the curated ontology's (Jaccard similarity 0.21--0.22).

The model that writes the ontologies matters. Claude Opus~5.5, DeepSeek-V4-Flash and Qwen3.8-27B each wrote three ontologies by the same recipe; on UniProt's artificial faults and averaged over each model's three ontologies, Opus's ontologies catch 60.4\% of the faults that change the answer and flag 10.3\% of the untouched correct queries, DeepSeek's catch 84.1\% and flag 59.5\%, and Qwen's catch 81.9\% and flag 61.0\%.
Whether a model's ontology misses its own agent's errors was also registered, as an interaction $I$ between two models predicted to be negative; the result is inconclusive: of the two pairs whose agents both give at least 20 wrong answers, the one on the masked copy of Rhea has $I=-9.2$ ($-20.1$ to $1.0$), and the one on UniProt, DeepSeek and Qwen3.8, has $I=+8.5$ ($1.5$ to $15.3$).

\subsection{The cost of checking}
\label{sec:app-cost}

Table~\ref{tab:cost} reports what checking costs on UniProt, per kind of ontology. Setup is incurred once per graph\revg{. The curated ontology is used as published, and the cost of authoring it is borne by its publisher; each ontology Claude Opus~5.5 wrote for UniProt cost under one US dollar at batch prices; the time to build the data-derived ontology and the time a check takes were not recorded.} Aligning the private graph's curated ontology with the data's identifiers took an expert three hours (Appendix~\ref{sec:app-graphs}). \revg{What differs is the retry after a flag: one after a curated flag uses more tokens and time than one after a data-derived flag, and the curated check's specific message costs about as much as a generic request to re-check.}

\begin{table}[!tb]
  \caption{What checking costs on UniProt, per kind of ontology\revg{, pooled over the seven models} (median, with the 90th percentile in parentheses).}
  \label{tab:cost}
  \scriptsize
  \setlength{\tabcolsep}{2.5pt}
  \renewcommand{\arraystretch}{0.9}
  \begin{tabular*}{\linewidth}{@{\extracolsep{\fill}}lcccc@{}}
    \toprule
    & Setup & Per answer checked & \multicolumn{2}{c}{Per answer flagged} \\
    \cmidrule(lr){4-5}
    Check from & \revg{(US\$)} & triples read & tokens & \revg{time (s)} \\
    \midrule
    \revg{Curated} & \revg{---} & \revg{25 (277)} & \revg{65,976 (435,066)} & \revg{46.0 (420.0)} \\
    \revg{\quad same flags, generic message} & \revg{---} & \revg{---} & \revg{60,927 (308,426)} & \revg{49.2 (309.7)} \\
    \revg{Data-derived} & \revg{---} & \revg{25 (269)} & \revg{28,076 (202,405)} & \revg{17.2 (112.3)} \\
    \revg{LLM-generated, draft 1} & \revg{0.43} & \revg{24 (267)} & \revg{42,027 (181,818)} & \revg{22.4 (170.5)} \\
    \revg{LLM-generated, draft 2} & \revg{0.47} & \revg{25 (269)} & \revg{58,719 (455,836)} & \revg{29.5 (456.1)} \\
    \revg{LLM-generated, draft 3} & \revg{0.39} & \revg{24 (267)} & \revg{65,030 (369,424)} & \revg{25.7 (342.7)} \\
    \revg{LLM-generated, masked (three)} & \revg{0.48--0.98} & \revg{25 (269)} & \revg{---} & \revg{---} \\
    \revg{Combined} & \revg{---} & \revg{50 (552)} & \revg{63,819 (429,171)} & \revg{28.7 (337.5)} \\
    \bottomrule
  \end{tabular*}
\end{table}

\section{Beyond the pooled comparison}
\label{sec:app-beyond}

This part asks whether the comparison behind RQ1 holds beyond the pooled UniProt results of Table~\ref{tab:main}. It repeats the comparison per model and under the registered sensitivities, against the comparators, on human-written questions, and on the other graphs.

\subsection{Per model and under the registered sensitivities}
\label{sec:app-models}
\label{sec:app-sensitivity}

Table~\ref{tab:app-models} gives each model's accuracy with each kind of ontology in the prompt, the change in accuracy the curated check makes through its retry, and the effect of the check's specific message per flagged answer. Table~\ref{tab:app-models-checks} gives each model's catch and false-alarm rates on its answers given with no ontology: for each of the six models with enough correct answers (all but Qwen3-Coder), the curated check flags more correct answers than the data-derived and masked LLM-generated ontologies and catches more wrong ones, so the pooled pattern of Table~\ref{tab:main} is not one model's. Sonnet and Haiku are wrong on fewer than 30 answers and Qwen3-Coder gives 44 answers, so their rates rest on few answers. The per-model rate of invented identifiers, registered for this appendix, was not computed.

\begin{table}[!tb]
  \caption{Per model, on UniProt: accuracy with each ontology in the prompt, in \%, and the curated check's net change after its retry and the specific message's effect per flagged answer (C5), in points.}
  \label{tab:app-models}
  \scriptsize
  \setlength{\tabcolsep}{1.5pt}
  \renewcommand{\arraystretch}{0.9}
  \begin{tabular*}{\linewidth}{@{\extracolsep{\fill}}lcccccc@{}}
    \toprule
    & \multicolumn{4}{c}{Accuracy, ontology in the prompt} & \multicolumn{2}{c}{Curated check} \\
    \cmidrule(lr){2-5}\cmidrule(lr){6-7}
    Model & none & curated & data-derived & LLM-generated & net change & message effect \\
    \midrule
    Claude Sonnet 5.5 & \rev{\resUniprotSonnetNonePromptAccuracy} & \rev{\resUniprotSonnetCuratedPromptAccuracy} & \rev{\resUniprotSonnetCensusPromptAccuracy} & \rev{\resUniprotSonnetGeneratedOnePromptAccuracy} & \rev{\resUniprotSonnetNoneCheckerCuratedNetChange} & \rev{\resConfCFiveSonnetEst} \\
    Claude Haiku 5.5 & \rev{\resUniprotClaudeHaikuFiveFiveNonePromptAccuracy} & \rev{\resUniprotClaudeHaikuFiveFiveCuratedPromptAccuracy} & \rev{\resUniprotClaudeHaikuFiveFiveCensusPromptAccuracy} & --- & \rev{\resUniprotClaudeHaikuFiveFiveNoneCheckerCuratedNetChange} & \rev{\resConfCFiveClaudeHaikuFiveFiveEst} \\
    Qwen3.8-27B & \rev{\resUniprotQwenNonePromptAccuracy} & \rev{\resUniprotQwenCuratedPromptAccuracy} & \rev{\resUniprotQwenCensusPromptAccuracy} & \rev{\resUniprotQwenGeneratedOnePromptAccuracy} & \rev{\resUniprotQwenNoneCheckerCuratedNetChange} & \rev{\resConfCFiveQwenEst} \\
    Qwen3-Coder-30B-A3B & \rev{\resUniprotQwenQwenThreeCoderThreeZeroBAThreeBInstructNonePromptAccuracy} & \rev{\resUniprotQwenQwenThreeCoderThreeZeroBAThreeBInstructCuratedPromptAccuracy} & \rev{\resUniprotQwenQwenThreeCoderThreeZeroBAThreeBInstructCensusPromptAccuracy} & --- & \rev{\resUniprotQwenQwenThreeCoderThreeZeroBAThreeBInstructNoneCheckerCuratedNetChange} & \rev{\resConfCFiveQwenQwenThreeCoderThreeZeroBAThreeBInstructEst} \\
    GLM-5.3-Flash & \rev{\resUniprotZaiOrgGLMFiveThreeFlashNonePromptAccuracy} & \rev{\resUniprotZaiOrgGLMFiveThreeFlashCuratedPromptAccuracy} & \rev{\resUniprotZaiOrgGLMFiveThreeFlashCensusPromptAccuracy} & --- & \rev{\resUniprotZaiOrgGLMFiveThreeFlashNoneCheckerCuratedNetChange} & \rev{\resConfCFiveZaiOrgGLMFiveThreeFlashEst} \\
    Gemma-4-31B & \rev{\resUniprotGoogleGemmaFourThreeOneBNonePromptAccuracy} & \rev{\resUniprotGoogleGemmaFourThreeOneBCuratedPromptAccuracy} & \rev{\resUniprotGoogleGemmaFourThreeOneBCensusPromptAccuracy} & --- & \rev{\resUniprotGoogleGemmaFourThreeOneBNoneCheckerCuratedNetChange} & \rev{\resConfCFiveGoogleGemmaFourThreeOneBEst} \\
    DeepSeek-V4-Flash & \rev{\resUniprotDeepseekAiDeepSeekVFourFlashZeroSevenThreeOneNonePromptAccuracy} & \rev{\resUniprotDeepseekAiDeepSeekVFourFlashZeroSevenThreeOneCuratedPromptAccuracy} & \rev{\resUniprotDeepseekAiDeepSeekVFourFlashZeroSevenThreeOneCensusPromptAccuracy} & --- & \rev{\resUniprotDeepseekAiDeepSeekVFourFlashZeroSevenThreeOneNoneCheckerCuratedNetChange} & \rev{\resConfCFiveDeepseekAiDeepSeekVFourFlashZeroSevenThreeOneEst} \\
    \bottomrule
  \end{tabular*}
\end{table}

\begin{table}[!tb]
  \caption{Each model's checks on UniProt, on its answers with no ontology in the prompt: catch / false alarm, in \%. The three masked LLM-generated ontologies give identical rates.}
  \label{tab:app-models-checks}
  \scriptsize
  \setlength{\tabcolsep}{3pt}
  \begin{tabular}{@{}lccc@{}}
    \toprule
    \revg{Model} & \revg{Curated} & \revg{Data-derived} & \revg{LLM-generated, masked} \\
    \midrule
    \revg{Claude Sonnet 5.5} & \revg{7.1 / 17.4} & \revg{0.0 / 1.7} & \revg{0.0 / 1.2} \\
    \revg{Claude Haiku 5.5} & \revg{14.8 / 15.9} & \revg{0.0 / 5.0} & \revg{0.0 / 2.2} \\
    \revg{Qwen3.8-27B} & \revg{13.2 / 16.7} & \revg{0.8 / 3.5} & \revg{0.0 / 2.0} \\
    \revg{Qwen3-Coder-30B-A3B} & \revg{32.4 / 30.0} & \revg{35.3 / 10.0} & \revg{14.7 / 0.0} \\
    \revg{GLM-5.3-Flash} & \revg{14.3 / 18.1} & \revg{1.9 / 5.6} & \revg{1.9 / 2.7} \\
    \revg{Gemma-4-31B} & \revg{30.2 / 19.2} & \revg{24.0 / 6.1} & \revg{16.7 / 5.1} \\
    \revg{DeepSeek-V4-Flash} & \revg{19.6 / 15.1} & \revg{4.5 / 1.4} & \revg{2.7 / 0.4} \\
    \bottomrule
  \end{tabular}
\end{table}

Re-running or excluding the attempts disturbed by time-outs (4.8\% of all attempts) leaves every registered decision unchanged and moves no estimate by more than 0.3 points; the data-derived ontology gives identical rates on real answers at frequency thresholds of 2 and 20 instead of 5; and no attempt was missing or refused. Under the earlier version of the scorer's rules (the frozen rules only add correct verdicts), C1a is $-1.2$ points ($-2.7$ to $0.2$), and excluding the two questions whose curated prompt contains a correct answer value leaves it at $-2.0$ ($-3.3$ to $-0.8$).  The unadjusted bootstrap $p$-values are \resConfCOneAPBootstrap{} (C1a), \resConfCOneBPBootstrap{} (C1b), \resConfCTwoPBootstrap{} (C2), \resConfCThreePBootstrap{} (C3), \resConfCFourPBootstrap{} (C4) and \resConfCFivePBootstrap{} (C5),, and a cluster-robust (CR2) Wald test agrees. Three registered source sensitivities and a registered selective check were not run.

\subsection{Comparators}
\label{sec:app-baselines}

Table~\ref{tab:app-baselines} places the ontology-derived checks beside the comparators of Section~\ref{sec:baselines}. The LLM judge reads the question, the query, the results and the answer; it is played by the agent's own model on real answers and by Claude Sonnet~5.5 and Haiku~5.5 on the 2,238 artificial faults whose question has a text. On real answers it catches more wrong answers than the curated or the data-derived ontology, at about the data-derived ontology's precision; on artificial faults it catches nearly every fault, including every change of meaning, and flags 30--38\% of the faulty queries whose answer is nonetheless correct. SPARQL-LLM's validator does no better than a random flag at its rate. The Ontology-based Query Check's eight rules together flag most answers, wrong and correct alike. Acted on, the judge's flags corrupted more correct answers than they fixed wrong ones, and the other checks' retries changed few answers. \revg{Combining the curated and data-derived checks adds 3 catches to the curated check's at 28 more false alarms, and 59 to the data-derived check's at 407 more.}

\begin{table}[!tb]
  \caption{Comparators on UniProt, in \%, on real answers and on artificial faults with a question text. Vs rand.: points by which precision beats a random flag at the same rate; fix / corr.: answers fixed and corrupted after one retry; the judge on faults: Sonnet / Haiku.}
  \label{tab:app-baselines}
  \scriptsize
  \setlength{\tabcolsep}{1.8pt}
  \renewcommand{\arraystretch}{0.9}
  \begin{tabular*}{\linewidth}{@{\extracolsep{\fill}}lcccccccc@{}}
    \toprule
    & \multicolumn{5}{c}{\rev{Real answers}} & \multicolumn{3}{c}{\rev{Artificial faults}} \\
    \cmidrule(lr){2-6}\cmidrule(lr){7-9}
    \rev{Check} & \rev{catch} & \rev{false al.} & \rev{prec.} & \rev{vs rand.} & \rev{\revg{fix / corr.}} & \rev{catch} & \rev{false al.} & \rev{prec.} \\
    \midrule
    \rev{Retry on empty} & \rev{2.5} & \rev{0.1} & \rev{81.3} & \rev{$+65.8$} & \rev{2 / 0} & \rev{80.4} & \rev{0.0} & \rev{100} \\
    \rev{Curated} & \rev{18.9} & \rev{16.8} & \rev{17.1} & \rev{$+1.6$} & \rev{28 / 5} & \rev{59.7} & \rev{13.2} & \rev{60.1} \\
    \rev{Data-derived} & \rev{8.2} & \rev{3.5} & \rev{29.9} & \rev{$+14.4$} & \revg{3 / 1} & \rev{75.5} & \rev{9.3} & \rev{73.0} \\
    \revg{Combined} & \revg{19.5} & \revg{17.8} & \revg{16.7} & \revg{$+1.3$} & \revg{30 / 5} & --- & --- & --- \\
    \rev{LLM judge} & \rev{40.3} & \rev{16.2} & \rev{31.3} & \rev{$+15.8$} & \rev{40 / 83} & \rev{98.7 / 98.0} & \rev{11.4 / 11.6} & \rev{74.2 / 73.8} \\
    \rev{SPARQL-LLM} & \rev{5.5} & \rev{6.5} & \rev{13.5} & \rev{$-2.0$} & \rev{2 / 1} & --- & --- & --- \\
    \rev{Query Check, 8 rules} & \rev{87.8} & \rev{90.2} & \rev{15.1} & \rev{$-0.4$} & --- & --- & --- & --- \\
    \rev{Query Check, 1--6} & \rev{34.8} & \rev{22.1} & \rev{22.3} & \rev{$+6.9$} & --- & --- & --- & --- \\
    \rev{Always flag} & \rev{100} & \rev{100} & \rev{15.5} & --- & --- & \rev{100} & \rev{100} & \rev{25.0} \\
    \rev{Never flag} & \rev{0.0} & \rev{0.0} & --- & --- & --- & \rev{0.0} & \rev{0.0} & --- \\
    \bottomrule
  \end{tabular*}
\end{table}

\subsection{Human-written questions}
\label{sec:app-human}

SIB's published human-written questions were also run with no ontology in the prompt, for Sonnet \revg{and Qwen3.8} on UniProt and Rhea (Table~\ref{tab:app-human}); they are analysed apart from the generated questions. Of SIB's 173 primary UniProt questions, 170 are analysed, since three ask for a graph, not an answer; of its 124 Rhea examples, 101, after excluding 18 with a SERVICE clause, by a rule fixed before the gold run, and 5 with an empty gold answer. The agents are far less accurate than on the generated questions, and most of their submitted answers are wrong. On the UniProt questions the LLM judge, the curated check and the data-derived ontology keep their order by catch rate, and \revg{Sonnet's} curated check beats a random flag at its rate (by 8.7 points, 1.9 to 15.5); on the Rhea questions it does no better than one. The false-alarm rates rest on \revg{few} correct answers, so they are not compared with those on the generated questions.

\begin{table}[!tb]
  \caption{Human-written questions (SIB) on UniProt and Rhea, no ontology in the prompt: accuracy and each check's catch / false alarm, in \%. Vs rand.: points by which precision beats a random flag; LLM-generated: range over three.}
  \label{tab:app-human}
  \scriptsize
  \setlength{\tabcolsep}{3pt}
  \begin{tabular}{@{}lcccc@{}}
    \toprule
    & \revg{Sonnet, UniProt} & \revg{Sonnet, Rhea} & \revg{Qwen3.8, Rhea} & \revg{Qwen3.8, UniProt} \\
    \midrule
    \revg{Questions} & \revg{170} & \revg{101} & \revg{101} & \revg{170} \\
    \revg{Accuracy} & \revg{11.8} & \revg{11.9} & \revg{9.2} & \revg{9.0} \\
    \revg{Curated} & \revg{34.3 / 11.7} & \revg{41.8 / 41.7} & \revg{35.3 / 46.4} & \revg{35.3 / 4.3} \\
    \revg{\quad vs rand.} & \revg{$+8.7$} & \revg{0.0} & \revg{$-3.2$} & \revg{$+10.7$} \\
    \revg{Data-derived} & \revg{0.8 / 0.0} & \revg{2.8 / 0.0} & \revg{1.0 / 0.0} & \revg{5.6 / 0.0} \\
    \revg{LLM-generated} & \revg{2.4--18.4 / 5.0--28.3} & \revg{3.3--6.6 / 8.3} & \revg{4.8--8.7 / 7.1} & \revg{0.9--18.0 / 6.5--19.6} \\
    \revg{LLM judge} & \revg{53.2 / 8.3} & \revg{29.6 / 0.0} & --- & --- \\
    \revg{\quad vs rand.} & \revg{$+11.5$} & \revg{$+14.5$} & --- & --- \\
    \bottomrule
  \end{tabular}
\end{table}

\begin{table*}[!t]
  \caption{The checks layer by layer on UniProt, caught / false alarms in \%: type swaps and untouched correct queries (artificial faults, per question), and real answers. C: class; DR: domain and range; +DJ: plus disjointness (J: the data-derived co-occurrence patterns); LLM-generated: mean of three.}
  \label{tab:app-layers}
  \scriptsize
  \setlength{\tabcolsep}{2.5pt}
  \renewcommand{\arraystretch}{0.9}
  \begin{tabular*}{\linewidth}{@{\extracolsep{\fill}}lcccccccc@{}}
    \toprule
    & \multicolumn{4}{c}{\revg{Artificial faults (caught / false alarms)}} & \multicolumn{4}{c}{\revg{Real answers (caught / false alarms)}} \\
    \cmidrule(lr){2-5}\cmidrule(lr){6-9}
    \revg{Ontology} & \revg{C} & \revg{C+DR} & \revg{+DJ (J)} & \revg{full} & \revg{C} & \revg{C+DR} & \revg{+DJ (J)} & \revg{full} \\
    \midrule
    \revg{Curated} & \revg{26.5 / 0.0} & \revg{41.3 / 7.1} & \revg{41.3 / 7.1} & \revg{56.3 / 15.7} & \revg{3.1 / 0.6} & \revg{8.0 / 8.8} & \revg{8.0 / 8.8} & \revg{18.9 / 16.8} \\
    \revg{Data-derived} & \revg{46.8 / 0.0} & \revg{73.7 / 0.0} & \revg{78.5 / 7.1} & \revg{78.5 / 7.1} & \revg{1.0 / 0.2} & \revg{1.0 / 0.3} & \revg{4.8 / 2.1} & \revg{8.2 / 3.5} \\
    \revg{LLM-generated, masked} & \revg{31.9 / 0.0} & \revg{50.4 / 0.0} & \revg{50.4 / 0.0} & \revg{50.4 / 0.0} & \revg{1.0 / 0.2} & \revg{1.0 / 0.3} & \revg{1.0 / 0.3} & \revg{5.0 / 1.8} \\
    \revg{LLM-generated, ordinary} & \revg{30.4 / 0.0} & \revg{46.8 / 0.0} & \revg{46.8 / 0.1} & \revg{55.3 / 10.3} & --- & --- & --- & --- \\
    \bottomrule
  \end{tabular*}
\end{table*}

\subsection{The other graphs}
\label{sec:app-graphs}

Table~\ref{tab:app-graphs} repeats the central comparison on the four exploratory graphs, for the two confirmatory models, on real answers. Rhea's curated ontology declares no domain, range or disjointness axioms, so only class and declaration checks apply to it, and as context every ontology on Rhea is truncated to the prompt budget. Some of Rhea's provisional artificial faults were exposed, so no artificial-fault result is reported for Rhea.

On Rhea the pattern of Table~\ref{tab:main} repeats, at a higher cost for the curated check: over the six models with enough correct answers (all but Qwen3-Coder), it flags 23.2--63.0\% of them, against 15.1--19.2\% on UniProt and more than any other ontology, while the data-derived ontology and the first two LLM-generated ones flag few.  The context difference seen on UniProt does not carry over: of the seven models, the data-derived ontology in the prompt gives the higher accuracy only for Gemma (83.0\% against 40.6\%) and Qwen3-Coder (16.7\% against 9.4\%), the two weakest; for GLM the two are nearly equal (87.9\% and 87.8\%).

On the masked copy of Rhea, the three LLM-generated ontologies, written from masked names, stay within a few points of one another and of the data-derived ontology, flagging few correct answers and catching few wrong ones for every model, while the curated check's rates vary widely across models. On the corporate graph, whose curated ontology declares domain and range axioms but no disjointness, no model is less accurate with either ontology in the prompt, and the curated check flags 20.5--25.9\% of the correct answers of the six models with enough of them (all but Qwen3-Coder), while the data-derived and LLM-generated ontologies flag almost none. The setting a developer is likely to deploy, a data-derived context with a curated check, gains a little on UniProt but less, nothing or a loss on the other graphs: \revg{pooled over the seven models on UniProt, one retry raises accuracy by 1.0 point;} one retry changes Sonnet's accuracy by at most 0.5 point on each of the four graphs where it was run, and Qwen3.8's by $+1.6$ points (0.3 to 3.0) on UniProt and by $-3.8$ ($-5.9$ to $-1.9$) on the corporate graph.

On the private graph, with no ontology in the prompt, Sonnet, Haiku, Qwen3.8, GLM and DeepSeek answer at least 92.9\% of the questions correctly, \revg{Gemma 59.2\%,} and Qwen3-Coder rarely submits an answer. The curated check flags \revg{56.3--72.8\% of the correct answers of the six models}, while the data-derived ontology and the three LLM-generated ones, which give identical results, flag almost none. \revg{Only GLM and Gemma give enough wrong answers to compare catch rates: the curated check catches 73.0\% and 74.0\% of them, and the other ontologies none of GLM's and 13.0\% of Gemma's.} An expert also aligned the curated ontology's terms with the data's identifiers\label{sec:app-align}, in three hours for 115 terms. On the answers given with no ontology in the prompt, alignment raised the share of correct answers the checks flag for each of the \revg{six models, from 56.3--72.8\% to 69.7--85.1\%}; the results with an ontology in the prompt are in the replication package. \revg{Leaving out the 20 questions that the triple store's known query faults can affect keeps every model's accuracy within its interval.}

\begin{table}[!tb]
  \caption{The central comparison on the exploratory graphs, real answers of Sonnet and Qwen3.8: accuracy, and each check's catch / false alarm, in \%. LLM-generated: one range on the masked copy; identical on the corporate and private graphs. $^\dagger$Very few wrong answers.}
  \label{tab:app-graphs}
  \scriptsize
  \setlength{\tabcolsep}{2.5pt}
  \renewcommand{\arraystretch}{0.9}
  \begin{tabular*}{\linewidth}{@{\extracolsep{\fill}}lccccc@{}}
    \toprule
    & & & & \multicolumn{2}{c}{LLM-generated} \\
    \cmidrule(lr){5-6}
    Graph & accuracy & Curated & Data-derived & drafts 1--2 & draft 3 \\
    \midrule
    \revg{Rhea, Sonnet} & \revg{96.0} & \revg{8.0 / 63.0} & \revg{0.0 / 2.1} & \revg{0.0 / 3.0--3.3} & \revg{36.0 / 9.8} \\
    \revg{Rhea, Qwen3.8} & \revg{92.1} & \revg{20.4 / 39.1} & \revg{0.0 / 3.3} & \revg{0.0 / 4.5--4.7} & \revg{16.3 / 11.4} \\
    \revg{Masked Rhea, Sonnet} & \revg{68.3} & \revg{58.8 / 57.0} & \revg{0.0 / 0.0} & \multicolumn{2}{c}{\revg{0.5--4.0 / 0.0--3.7}} \\
    \revg{Masked Rhea, Qwen3.8} & \revg{30.0} & \revg{25.9 / 20.6} & \revg{1.9 / 2.6} & \multicolumn{2}{c}{\revg{0.9--2.8 / 2.6--3.7}} \\
    \revg{Corporate, Sonnet} & \revg{99.4} & \revg{50.0$^\dagger$ / 24.8} & \revg{0.0$^\dagger$ / 0.5} & \multicolumn{2}{c}{\revg{0.0$^\dagger$ / 0.0}} \\
    \revg{Corporate, Qwen3.8} & \revg{94.3} & \revg{22.2 / 25.9} & \revg{2.8 / 1.3} & \multicolumn{2}{c}{\revg{2.8 / 1.3}} \\
    \revg{Private, Sonnet} & \revg{100.0} & \revg{--- / 71.1} & \revg{--- / 1.6} & \multicolumn{2}{c}{\revg{--- / 1.6}} \\
    \revg{Private, Qwen3.8} & \revg{98.1} & \revg{50.0$^\dagger$ / 70.1} & \revg{0.0$^\dagger$ / 0.5} & \multicolumn{2}{c}{\revg{0.0$^\dagger$ / 0.5}} \\
    \bottomrule
  \end{tabular*}
\end{table}

\section{Where false alarms come from}
\label{sec:app-why}

This part collects the evidence behind RQ2 and behind the masked ontologies of C3. It switches the checks on layer by layer, describes the audit of false alarms and misses, and measures how far agents and LLMs rely on recognising a graph's vocabulary.

\begin{table}[!tb]
  \caption{Closed-book probes, exact match in \%: an answer (a) and a query (b) written from memory; and the agents' accuracy with no ontology, masked Rhea minus Rhea, in points with 95\% intervals.}
  \label{tab:app-closed}
  \scriptsize
  \setlength{\tabcolsep}{3pt}
  \begin{tabular}{@{}lcccccc@{}}
    \toprule
    & \revg{a: UniProt} & \multicolumn{4}{c}{\revg{b: query from memory}} & \revg{agent accuracy} \\
    \cmidrule(lr){3-6}
    \revg{Model} & & \revg{UniProt} & \revg{Rhea} & \revg{masked Rhea} & \revg{corporate} & \revg{masked $-$ Rhea} \\
    \midrule
    \revg{Sonnet} & \revg{4.9} & \revg{53.7} & \revg{46.0} & \revg{1.1} & \revg{46.3} & \revg{$-27.8$ ($-32.6$, $-22.9$)} \\
    \revg{Haiku} & \revg{1.4} & \revg{39.0} & \revg{24.8} & \revg{1.3} & \revg{0.0} & \revg{$-35.2$ ($-40.6$, $-29.8$)} \\
    \revg{Qwen3.8} & \revg{0.8} & \revg{0.0} & \revg{4.9} & \revg{0.0} & \revg{26.7} & \revg{$-62.1$ ($-68.2$, $-55.8$)} \\
    \revg{Qwen3-Coder} & \revg{0.0} & \revg{1.0} & \revg{0.0} & \revg{1.1} & \revg{0.0} & \revg{$-2.1$ ($-3.4$, $-0.9$)} \\
    \revg{GLM} & \revg{1.7} & \revg{26.7} & \revg{21.7} & \revg{0.0} & \revg{37.5} & \revg{$-52.9$ ($-58.0$, $-47.6$)} \\
    \revg{Gemma} & \revg{0.0} & \revg{0.6} & \revg{0.0} & \revg{1.0} & \revg{0.0} & \revg{$-26.5$ ($-32.3$, $-21.0$)} \\
    \revg{DeepSeek} & \revg{0.2} & \revg{1.9} & \revg{1.4} & \revg{0.6} & \revg{0.0} & \revg{$-50.5$ ($-56.2$, $-44.8$)} \\
    \bottomrule
  \end{tabular}
\end{table}

\subsection{The checks layer by layer}
\label{sec:app-layers}

Table~\ref{tab:app-layers} switches each ontology's checks on one layer at a time: class checks with the heuristic of Section~\ref{sec:check-types}, then domain and range, then disjointness (for the data-derived ontology, its checks on co-occurrence patterns instead), then the declaration check, which completes the full set. On real answers, the precision of the curated ontology's domain and range flags differs from that of a random flag at the same rate by $-1.2$ points (95\% CI $-5.5$ to $5.1$), and that of its class membership and term-declaration flags by $+8.0$ points (1.8 to 15.3). In the data-derived ontology, the class, domain and range axioms catch more type swaps than the same axioms in the masked ontologies; without the heuristic they catch 37.2\% of type swaps, against 73.4\% with it, pooled over items, and most of the data-derived ontology's false alarms arise from its checks on co-occurrence patterns.

\subsection{The audit}
\label{sec:app-audit}

\revg{Two annotators coded a seeded, stratified sample of 145 answers from the test questions, one answer per question within each stratum: 64 wrong answers that no check flagged and 32 correct answers that the curated check flagged, spread evenly over the four public graphs (the missed answers also over Claude and the open-weight models), 16 wrong answers given with the curated ontology in the prompt that the curated check passed, ten of each of the first two kinds on the private graph, and 13 items allocated by a fixed rule. Annotators saw the question, the correct and the agent's query and answer, and the check's message or the ontology where relevant, but not the model or the condition. They coded each missed answer's kind of error, each false alarm's cause, and whether a wrong answer of the third kind follows from a wrong or misleading statement in the ontology. Cohen's $\kappa$ over the 112 main items is 0.54 (kind of error), 0.79 (structural detectability), 0.77 (cause of a false alarm) and 0.33 (whether a wrong answer follows from the ontology); Table~\ref{tab:why} counts the codes both annotators gave.  Few wrong answers of the third kind followed from a statement in the ontology. On the private graph, every audited false alarm came from an undeclared term, and no missed answer could be flagged by a structural check.}

\subsection{Recognising a graph's vocabulary}
\label{sec:app-recall}

Each model was also asked, with no tools, to answer each test question from memory (probe a) or to write its query from memory, run on the graph (probe b, public graphs only; GLM on 40 questions per graph); the registered expectations were near zero for answers everywhere, and for queries clearly above zero on UniProt and Rhea and near zero on masked graphs (Table~\ref{tab:app-closed}). Answers from memory are exact for at most 9.8\% of the questions on any public graph and 1.1\% on the private graph. Queries from memory are exact for many questions on UniProt and Rhea for Sonnet, Haiku and GLM, and for almost none on the masked copy of Rhea, for every model. The other four models score low, but many of their replies contained no usable query, so their low scores do not by themselves show an absence of recall. The agents' own accuracy also falls on the masked copy, for every model and prompt. Probes on the human-written questions were registered but not run.